%% file: anonymous-submission-latex-2024.tex
\documentclass[letterpaper]{article} 
\usepackage{aaai24}  
\usepackage{times}  
\usepackage{helvet}  
\usepackage{courier}  
\usepackage[hyphens]{url}  
\usepackage{graphicx} 
\usepackage{natbib}  
\usepackage{caption} 
\usepackage{algorithm}
\usepackage{algorithmic}

\usepackage{newfloat}
\usepackage{listings}
\DeclareCaptionStyle{ruled}{labelfont=normalfont,labelsep=colon,strut=off} 
\floatstyle{ruled}
\newfloat{listing}{tb}{lst}{}
\floatname{listing}{Listing}
\input{math_commands.tex}

\usepackage{mathtools}
\usepackage{amssymb}
\usepackage{amsthm}
\usepackage[capitalise]{cleveref}
\usepackage{subcaption}
\usepackage{cancel}
\usepackage{booktabs}
\usepackage{siunitx}

\usepackage{comment}

\newcommand{\pg}{P\'{o}lya-Gamma }
\newcommand{\ind}{\boldsymbol{\mathrm{i}}}

\newcommand{\vect}{\mathrm{vec}}
\newcommand{\tr}{\mathrm{tr}}

\newcommand{\dif}{\mathop{}\!\mathrm{d}}

\newcommand{\ie}{\emph{i.e.}}
\newcommand{\eg}{\emph{e.g.}}
\newcommand{\wrt}{\emph{w.r.t.}}

\newcommand{\bftab}{\fontseries{b}\selectfont}

\theoremstyle{definition}
\newtheorem{definition}{Definition}

\usepackage{eqparbox}
\usepackage{xcolor}
\definecolor{flamingopink}{rgb}{0.99, 0.56, 0.67}

\title{Efficient Nonparametric Tensor Decomposition for Binary and Count Data}
\author {
    Zerui Tao\textsuperscript{\rm 1,2},
    Toshihisa Tanaka\textsuperscript{\rm 1,2},
    Qibin Zhao\textsuperscript{\rm 2,1}\thanks{Corresponding author}
}
\affiliations {
    \textsuperscript{\rm 1}Tokyo University of Agriculture and Technology, Japan\\
    \textsuperscript{\rm 2}RIKEN Center for Advanced Intelligence Project (AIP), Japan\\
    \texttt{zerui.tao@riken.jp},\, \texttt{tanakat@cc.tuat.ac.jp},\, \texttt{qibin.zhao@riken.jp}
}

\usepackage{bibentry}

\begin{document}

\maketitle

\begin{abstract}
  In numerous applications, binary reactions or event counts are observed and stored within high-order tensors.
  Tensor decompositions (TDs) serve as a powerful tool to handle such high-dimensional and sparse data.
  However, many traditional TDs are explicitly or implicitly designed based on the Gaussian distribution, which is unsuitable for discrete data.
  Moreover, most TDs rely on predefined multi-linear structures, such as CP and Tucker formats.
  Therefore, they may not be effective enough to handle complex real-world datasets.
  To address these issues, we propose ENTED,
  an \underline{E}fficient \underline{N}onparametric
  \underline{TE}nsor \underline{D}ecomposition for binary and count tensors.
  Specifically, we first employ a nonparametric Gaussian process (GP) to replace traditional multi-linear structures.
  Next, we utilize the \pg augmentation which provides a unified framework to establish conjugate models for binary and count distributions.
  Finally, to address the computational issue of GPs, we enhance the model by incorporating sparse orthogonal variational inference of inducing points, which offers a more effective covariance approximation within GPs and stochastic natural gradient updates for nonparametric models.
  We evaluate our model on several real-world tensor completion tasks, considering binary and count datasets.
  The results manifest both better performance and computational advantages of the proposed model.
\end{abstract}

\section{Introduction}%
\label{sec:intro}

Tensor data are ubiquitous in many real-world applications,
such as visual processing \citep{liu2012tensor,zhao2015bayesian},
spatial-temporal forecasting \citep{NIPS2014_aa2a7737,pmlr-v130-qiu21a},
probabilistic modeling \citep{glasser2019expressive,novikov2021tensor},
among many others.
Tensor decomposition (TD) is a powerful tool for handling such high-order data.
Due to the large tensor sizes, TD aims to factorize the original data into
much smaller tensor factors.
Through these sharing tensor factors,
underlying structures or correlations among different tensor modes can be captured.
Based on this idea, many elegant TD models have been proposed,
such as CP decomposition \citep{hitchcock1927expression}, Tucker decomposition \citep{tucker1966some}, tensor train/ring decomposition \citep{oseledets2011tensor,zhao2019learning}
and many variants \citep{kolda2009tensor,cichocki2016tensor}.
Choosing proper TD structures often involves domain knowledge and
can largely affect the final performance \citep{li2020evolutionary,li2022permutation}.

While most TD models focus on continuous problems,
many real-world applications may encounter binary or count data.
For example, in click-trough-rate (CTR) prediction tasks,
a tensor of shape \emph{user} \(\times\) \emph{item} \(\times\) \emph{time}
may store records of whether a specific user clicked on the item at the time.
Additionally, many tensors consist of multi-way events,
where each element is the count of the event history \citep{schein2015bayesian,schein2016bayesian}.
For example, the Covid-19 dataset \citep{dong2020interactive} contains the
number of infection claims, and each observation is associated with several
attributes (tensor modes) such as locations, time, and claim types.
However, less effort is made to deal with such discrete observations.
Compared to continuous counterparts,
handling discrete distributions brings additional difficulties when constructing probabilistic models,
due to their non-differentiable and non-conjugacy nature.

To address these issues, we propose ENTED,
an \underline{E}fficient \underline{N}onparametric
\underline{TE}nsor \underline{D}ecomposition for binary and count data.
The proposed model is inherited from the Gaussian process tensor factorization \citep[GPTF,][]{zhe2016distributed}.
Specifically, we adopt a Gaussian process (GP) to replace traditional
multi-linear contraction rules.
The great flexibility of nonparametric GPs enables us to
learn underlying structures of complex real-world datasets adaptively,
rather than picking one beforehand.
To cope with discrete data,
the \pg (PG) augmentation \citep{polson2013bayesian} is adopted,
which provides a unified framework to establish conjugate models
for both binary and count data \citep{klami2015polya}.
Moreover, to efficiently approximate infeasible covariance matrices
in GPs, we derive a novel sparse orthogonal variational inference
\citep[SOLVE,][]{shi2020sparse} scheme for our model that incorporates
PG augmentation and natural gradient (NG) updates.
Notably, our model allows for stochastic
optimization that is scalable to large tensors.
The contributions are summarized as follows:
\begin{itemize}
  \item We propose a flexible nonparametric TD for binary and count data.
        By using GPs, the model can adaptively learn complex hidden structures
        of high-order tensors.
  \item A PG augmentation scheme is adopted, resulting in a unified augmented model for binary and count data. Due to the conjugacy, efficient NG
        updates can be derived.
  \item To obtain an efficient covariance approximation,
        we derive a SOLVE framework with PG augmentation and NG updates for
        nonparametric tensor factorization,
        which enables fast and stochastic optimization.
\end{itemize}

Finally, we demonstrate the proposed model on binary and count tensor completion tasks.
Our model shows superior prediction accuracy and distributional estimation
on the six real-world datasets.
In addition, ablation studies on inducing points are conducted to show
the effectiveness and computational benefits of our model.

\section{Backgrounds}%
\label{sec:backgrounds}

\subsection{Notations}

We denote scalars, vectors, matrices, and tensors as
lowercase letters, bold lowercase letters, bold capital letters, and sans-serif bold capital letters, \eg,
$x, \vecs{x}, \mX$ and $\tX$, respectively.
For an order-$D$ tensor $\tX \in \mathbb{R}^{I_1 \times \cdots \times I_D}$, we denote its $(i_1, \dots, i_{D})$-th entry
as $x_{\ind}$, where $\ind = (i_1, \dots, i_{D})$.
Moreover,
$\mathcal{N}(\cdot, \cdot)$ denotes Normal distribution,
$\mathcal{B}(\cdot)$ denotes Bernoulli distribution,
$NB(\cdot, \cdot)$ denotes negative binomial (NB) distribution,
$PG(\cdot)$ denotes \pg (PG) distribution
and $\KL(p \lVert q)$ denotes the Kullback-Leibler (KL) divergence between two distributions $p$ and $q$.

\subsection{Tensor Decomposition}

Tensor decomposition \citep[TD,][]{kolda2009tensor} aims to factorize an order-$D$ tensor $\tX \in \mathbb{R}^{I_1 \times \cdots \times I_D}$ into $D$ smaller
latent factors $\mZ^{(d)} \in \mathbb{R}^{I_d \times R_d}, \forall d = 1, \dots, D$, where the sequence $(R_1, \dots, R_{d})$ is the tensor rank.
Traditional TDs depend on predefined contraction rules. For example, the CP decomposition \citep{hitchcock1927expression} assumes,
\begin{equation}
\label{eq:intro-cp}
  x_{\ind} = \sum_{r=1}^R \lambda_r z^{(1)}_{i_1 r} \cdots z^{(D)}_{i_D r},
\end{equation}
where $\lambda_r$ are factor weights for each rank-1 components and $R = R_1 = \cdots = R_D$.
Tucker decomposition \citep{tucker1966some} extends CP to have multiway weights, \ie,
$\displaystyle x_{\ind} = \sum_{r_{1}=1}^{R_{1}} \cdots \sum_{r_{D}=1}^{R_{D}} \lambda_{r_1 \dots r_{D}} z^{(1)}_{i_1 r_{1}} \cdots z^{(D)}_{i_D r_{D}}.$
Other popular TDs include tensor train \citep[TT,][]{oseledets2011tensor}, tensor ring \citep[TR,][]{zhao2019learning}, t-SVD \citep{kilmer2013third} and many variants \citep{kolda2009tensor,cichocki2016tensor}.
All these TD methods employ predefined multi-linear structures, possibly insufficient to cope with complex real-world datasets.

To mitigate the issue, a line of work \citep{pmlr-v5-chu09a,xu2012infinite,zhe2016distributed} studied nonparametric TDs, which can adaptively learn nonlinear structures from data.
Here, we briefly introduce the Gaussian process tensor factorization \citep[GPTF,][]{zhe2016distributed}
due to its flexibility and scalability.
In particular, given latent factors $\mZ^{(d)}\in \mathbb{R}^{I_d \times R}, \forall d=1, \dots, D$ with $R = R_1 = \cdots = R_D$,
we denote latent factors associated with index $\ind$ as
$\vm_{\ind} = [\vz^{(1)}_{i_{1}}, \dots, \vz^{(D)}_{i_{D}}] \in \mathbb{R}^{DR}$,
where $\vz^{(d)}_{i_{d}} \in \mathbb{R}^{R}$ is the $i_{d}$-th row of $\mZ^{(d)} $.
In GPTF,
the linear contraction forms of traditional TDs (\eg, \cref{eq:intro-cp})
is characterized by a GP,
\begin{equation*}
p(x_{\ind}) = \mathcal{N}(x_{\ind} \mid f_{\ind}, \beta^{-1}), \, p(\vf) = \mathcal{N}(\vf \mid \vecs{0}, k(\mM_{\Omega}, \mM_{\Omega})),
\end{equation*}
where $k(\cdot, \cdot)$ is a kernel function, $\Omega = \{ \ind_1, \dots, \ind_{N}  \}$ denotes all observed indices and $\mM_{\Omega} \in \mathbb{R}^{N \times DR}$ is concatenated by latent factors associating with all observed entries, \ie, $\{ \vm_{\ind}: \ind \in \Omega \}$.
To deal with binary data, \citet{zhe2016distributed} adopted an augmented variable
$\omega_n, \forall n = 1, \dots, N$ and the Probit model,
\begin{equation*}
p(x_{\ind_{n}} \mid \omega_{n}) = \mathcal{B}(\Phi(\omega_{n})), \quad p(\omega_n \mid f_{n}) = \mathcal{N}(\omega_n \mid f_n, 1),
\end{equation*}
where $\Phi(\cdot)$ is the cumulative distribution function (CDF) of standard Normal distribution.
Finally, the joint distribution for binary GPTF becomes,
\begin{equation}\label{eq:gptf-origin}
\begin{multlined}[.4\textwidth]
  p(\vx_{\Omega}, \vf, \vecs{\omega}, \mZ) =  p(\vf \mid \mZ) \cdot \prod_{d=1}^{D} p(\mZ^{(d)}) \cdot \\
  \prod_{n=1}^N p(x_{\ind_n} \mid \omega_{n}) p(\omega_{n} \mid f_{n}),
\end{multlined}
\end{equation}
where we denote $\mZ = \{ \mZ^{(d)}\}_{d=1}^{D}$ for simplicity.

However, learning this model requires cubic complexity with sample sizes, \ie, $\mathcal{O}(N^{3})$, which is prohibited in real applications with massive observations.
To address the issue, \citet{zhe2016distributed} adopted the sparse variational GP \citep[SVGP,][]{titsias2009variational} framework
and derived a distributed evidence lower bound (ELBO) analogous to \citet{gal2014distributed}.
In specific, a small set of inducing inputs $\mB \in \mathbb{R}^{p \times DR}$ and points $\vu \in \mathbb{R}^{p}$ are introduced.
By assuming the observations are conditionally independent given the inducing points,
the probabilistic model in \cref{eq:gptf-origin} can be expressed as
\begin{equation}\label{eq:gptf-sto}
\begin{multlined}[.4\linewidth]
  p(\vx_{\Omega}, \vf, \vecs{\omega}, \mZ, \vu) = \prod_{d=1}^{D} p(\mZ^{(d)}) \cdot \\
  \prod_{n=1}^N p(x_{\ind_n} \mid \omega_{n}) p(\omega_{n} \mid f_{n}) \cdot p(\vf \mid \vu, \mZ) p(\vu),
\end{multlined}
\end{equation}
where
\begin{align}
  p(\vf \mid \vu, \mZ) &= \mathcal{N}(\vf \mid \mK_{MB}\mK_{BB}^{-1} \vu, \tilde{\mK}), \label{eq:gp-fu} \\
  p(\vu) &= \mathcal{N}(\vu \mid \vzero, \mK_{BB}). \label{eq:gp-u}
\end{align}
For simplicity, we denote $\mK_{MB} = k(\mM_{\Omega}, \mB), \mK_{BB} = k(\mB, \mB), \mK_{MM} = k(\mM_{\Omega}, \mM_{\Omega})$ and $\tilde{\mK} = \mK_{MM} - \mK_{MB} \mK_{BB}^{-1} \mK_{BM}$.
To learn posteriors of the inducing points $\vu$ and the augmented variable $\vecs{\omega}$, a variational distribution $q(\vu, \vecs{\omega}) = q(\vu) q(\vecs{\omega})$
is adopted. The ELBO becomes,
\begin{equation}\label{eq:gptf-elbo}
\begin{multlined}[.4\linewidth]
  \log p(\vx_{\Omega}, \mZ) \geq \\
    \E_{p(\vf \mid \vu) q(\vu) q(\vecs{\omega})} \log p(\vx_{\Omega} \mid \mZ, \vf, \vu, \vecs{\omega}) - \\
    \KL(q(\vu) \lVert p(\vu)) - \KL(q(\vecs{\omega}) \lVert p(\vecs{\omega})) + \log p(\mZ).
\end{multlined}
\end{equation}
\citet{zhe2016distributed} showed that analytical solutions of $q(\vu), q(\vecs{\omega})$
can be obtained and the collapsed form of \cref{eq:gptf-elbo} can be computed in a \emph{distributed} manner.
The computational complexity is reduced to $\mathcal{O}(N p^2 + p^3)$.
Nevertheless, this approach can not handle integer observations like count data.
Also, the distributed objective cannot be optimized in a stochastic way,
which is desirable in many applications, \eg, when powerful computing clusters are not available or
the samples come in streams.
To enable stochastic optimization, we present an extension of GPTF in \cref{sec:app-gptf}.

\section{Proposed Model}%
\label{sec:model}

\subsection{Nonparametric Tensor Decomposition with~\pg Augmentation}\label{sec:pg-augment}

While many tensor data encounter discrete observations,
GPTF \citep{zhe2016distributed} cannot deal with them directly,
since discrete distributions yield non-conjugate GPs.
To address this issue,
we improve GPTF by employing the PG augmentation \citep{polson2013bayesian},
that provides a unified framework to establish conjugate 
models for Bernoulli and NB distributions \citep{klami2015polya}.

\subsubsection{\pg Augmentation}

A PG variable $\omega \sim PG( b, c)$ is defined as \citep{polson2013bayesian},
  \[ \omega \stackrel{D}{=} \frac{1}{2 \pi^{2}} \sum_{k=1}^{\infty} \frac{g_k}{(k - 1 / 2)^2 + c^2 / (4 \pi^{2})}, \]
where $g_k \overset{\mathrm{iid}}{\sim} Ga(b, 1)$,
and $\stackrel{D}{=}$ means equality in distribution.
Given PG variable $\omega \sim PG(b, 0)$, we have,
\begin{equation}\label{eq:pg-exp-exp}
\begin{multlined}[.7\linewidth]
\frac{\exp(t)^a}{(1 + \exp(t))^b} = \\ 2^{-b} \exp((a - b/2) t) \int_0^{\infty} \exp(- \omega t^2 / 2) p(\omega) \dif \omega.
\end{multlined}
\end{equation}

For binary data $x_{\ind} \in \{ 0, 1\}, \forall \ind \in \Omega$,
we adopt the logistic transform,
\begin{equation*}
  p(x_{\ind_n} \mid f_{n}) = \mathcal{B}(x_{\ind_n} \mid \sigma(f_{n})) = (1 + \exp(- f_{n}))^{-1},
\end{equation*}
where $\sigma(\cdot)$ is the logistic function. For count data
$x_{\ind} \in \mathbb{N}, \forall \ind \in \Omega$, we adopt
the negative binomial (NB) model,
\begin{equation*}
  p(x_{\ind_n} \mid f_{n}) = NB(x_{\ind_n} \mid \zeta, p_{n}) = \frac{\Gamma(\zeta + x_{\ind_n})}{x_{\ind_n}! \Gamma(\zeta)} p_{n}^{x_{\ind_n}} (1 - p_{n})^\zeta
\end{equation*}
where $p_{n} = 1 / (1 + \exp( - f_{n} ))$ and \emph{the number of successes} $\zeta$ is a hyper-parameter.
It was shown that the Bernoulli and NB distributions can be augmented
by PG variables as follows \citep{klami2015polya},
\begin{equation} \label{eq:pga}
  p(x_{\ind_n}, \omega_{n} \mid f_{n}) \propto 2^{-b} \exp( \chi_{n} f_{n} - \frac{1}{2} \omega_{n} f_{n}^{2} ) PG(\omega_{n} \mid b, 0) ,
\end{equation}
where $\omega_{n}$ is the PG augmented variable.
For binary data, $b = 1$ and $\chi_{n} = x_{\ind_n} - 1 / 2$.
For count data, $b = x_{\ind} + \zeta$ and $\chi_{n} = (x_{\ind_n} - \zeta) / 2$.
According to \cref{eq:pg-exp-exp}, the original distribution $p(\vx_{\Omega} \mid \vf)$ can be recovered
by marginalizing out the augmented variable $\omega_{n}$ in \cref{eq:pga}.
Note that \cref{eq:pga} admits a quadratic form, which is conjugate to Gaussian distribution.
Therefore, PG augmentation is widely used to handle non-Gaussian likelihoods.
Finally, by adopting the PG augmentation, the joint distribution of \cref{eq:gptf-sto} becomes,
\begin{equation}\label{eq:pg-joint}
  \begin{multlined}[.4\textwidth]
   p(\vx_{\Omega}, \vf, \vecs{\omega}, \mZ, \vu) =  \prod_{d=1}^{D} p(\mZ^{(d)}) \cdot \\
   p(\vx_{\Omega} \mid \vecs{\omega}, \vf) p(\vecs{\omega}) p(\vf \mid \vu, \mZ) p(\vu).
  \end{multlined}
\end{equation}
The prior $p(\mZ^{(d)})$ is chosen to be standard Gaussian.
Other distributions can be found in \cref{eq:gp-fu,eq:gp-u,eq:pga}.

\subsubsection{Evidence Lower Bound}

To marginalize out latent variables $\vf, \vecs{\omega}, \vu$ in the joint PDF \cref{eq:pg-joint} in a scalable way,
we have to seek for variational approximation.
In particular, we can assign $q(\vf, \vecs{\omega}, \vu) = p(\vf \mid \vu) q(\vecs{\omega}) q(\vu)$,
where
\begin{equation}\label{eq:q_omega_u}
q(\vecs{\omega}) = \prod_{n=1}^{N} PG(b, c_{n}), \quad q(\vu) = \mathcal{N}(\vmu^{(u)}, \mat{\Sigma}^{(u)}).
\end{equation}
Then, the ELBO becomes,
\begin{equation}\label{eq:svi-elbo1}
  \begin{multlined}[.4\textwidth]
    \log p_{\theta}(\vx_{\Omega}, \mZ) \geq \E_{p(\vf \mid \vu) q(\vecs{\omega}) q(\vu))} \log p(\vx_{\Omega} \mid \vecs{\omega}, \vf) - \\
    \KL( q(\vu)q(\vecs{\omega}) \lVert p(\vu)p(\vecs{\omega}) ) + \log p(\mZ).
  \end{multlined}
\end{equation}
The optimal solution for $q(\vecs{\omega})$ can be derived analytically, as we will show later.
The time complexity of \cref{eq:svi-elbo1} is $\mathcal{O}(N p^2 + p^3)$,
where $p$ is the number of inducing points. More importantly,
unlike in \citet{zhe2016distributed}, this objective \cref{eq:svi-elbo1} is factorized over samples. Hence, $N$ can be replaced by mini-batches, making it scalable to large datasets.

\subsubsection{Inference with Natural Gradients}

Maximizing the ELBO \cref{eq:svi-elbo1} yields efficient stochastic variational inference (SVI) with natural gradient (NG) updates.
Firstly, for local parameter $\vc$, we have the following optimal solution,
\begin{equation*}
c_n = \sqrt{ \tilde{k}_{n, n} + \vecs{\kappa}^{(u)}_n \mat{\Sigma}^{(u)} \vecs{\kappa}_n^{(u), \intercal} + \vmu^{(u), \intercal} \vecs{\kappa}_n^{(u)} \vecs{\kappa}^{(u), \intercal}_n \vmu^{(u)} },
\end{equation*}
where $\tilde{k}_{n, n}$ are diagonal elements of $\tilde{\mK}$, $\vecs{\kappa}^{(u)}_n $ is the $n$-th row of $\mK_{MB} \mK^{-1}_{BB}$ (we treat it as a column vector for notation consistency).
Then, we can derive the NGs of
natural parameters $\vecs{\eta}^{(u)}_{1} = \mat{\Sigma}^{(u), -1} \vmu^{(u)}$ and $\vecs{\eta}^{(u)}_{2} = - \frac{1}{2} \mat{\Sigma}^{(u), -1}$ as
\begin{align*}
\tilde{\triangledown}_{\vecs{\eta}_{1}^{(u)}} &= \frac{N}{s} \sum_{n \in \mathcal{S}}   \chi_{n}\vecs{\kappa}^{(u)}_{n} - \vecs{\eta}^{(u)}_{1}\\
\tilde{\triangledown}_{\vecs{\eta}_{2}^{(u)}} &= - \frac{1}{2} \left( \mK_{BB}^{-1} + \frac{N}{s} \sum_{n\in \mathcal{S}} \theta_{n} \vecs{\kappa}^{(u)}_{n} \vecs{\kappa}_{n}^{(u), \intercal} \right) - \vecs{\eta}^{(u)}_{2},
\end{align*}
where $\tilde{\triangledown}$ means NG. $\mathcal{S}$ denotes the set of mini-batch data with $s = | \mathcal{S} |$,
and $\displaystyle \theta_{n} = \frac{\tanh(c_{n} / 2)}{2 c_{n}}$.
Finally,
the latent factors $\mZ$ and inducing inputs $\mB$ are then optimized
by maximizing the ELBO \cref{eq:svi-elbo1} using gradient-based methods
such as Adam \citep{kingma2014adam}.
For detailed derivations, please check \cref{sec:app-elbo-ng}.

\subsection{Efficient Orthogonally Decoupled Approximation}\label{sec:solve}

Although the sparse variational GP tensor decomposition
presented in \cref{sec:pg-augment}
can handle large-scale binary or count tensors.
One may need to use many inducing points to approximate the full covariance matrix, which leads to large computational costs.
To allow efficient sparse approximation,
we derive another lower bound using the
sparse orthogonal variational inference \citep[SOLVE,][]{shi2020sparse}
framework for our model.

The idea is to decompose the GP into two orthogonal processes in the
reproduce Hilbert kernel space (RHKS), and then adopt two sets of inducing
points $\vu, \vv$ to approximate these two processes respectively.
Specifically, we decompose function $\vf$ in \cref{eq:gp-fu} into two orthogonal components,
\begin{equation*}
p(\vf_{\bot}) =  \mathcal{N}(\vzero, \tilde{\mK}), \quad \vf = \vf_{\bot} + \mK_{MB} \mK^{-1}_{BB} \vu,
\end{equation*}
where $\tilde{\mK}$ is defined in \cref{eq:gp-fu}.
Then, apart from inducing points $\vu$, we use another set of inducing points
$\vv$ to approximate $\vf_{\bot}$ separately,
\begin{equation*}
p(\vv) = \mathcal{N}(\vzero, \mK_{HH}),
\end{equation*}
where $\mK_{HH}$ is the covariance matrix of corresponding inducing inputs $\mH$.
The joint probability in \cref{eq:pg-joint} becomes,
\begin{equation}\label{eq:solve-joint}
\begin{multlined}[.8\linewidth]
  p(\vx_{\Omega}, \vf_{\bot}, \vecs{\omega}, \mZ, \vu, \vv) =  \prod_{d=1}^{D} p(\mZ^{(d)}) \cdot \\
   p(\vx_{\Omega} \mid \vecs{\omega}, \vf) p(\vecs{\omega}) p(\vu) p(\vf_{\bot} \mid \vv) p(\vv).
\end{multlined}
\end{equation}
Similar to \cref{eq:gp-fu,eq:gp-u}, we have
\begin{align*}
  p(\vf_{\bot} \mid \vv) &= \mathcal{N}(\mK_{MH}\mK_{HH}^{-1} \vv, \tilde{\mK} - \mK_{MH} \mK^{-1}_{HH} \mK_{HM}), \\
  p(\vv) &= \mathcal{N}(\vzero, \mK_{HH}),
\end{align*}
where $\mK_{MH} = k(\mM_{\Omega}, \mH)$ and $\mK_{HH} = k(\mH, \mH)$.

To get the variational lower bound, apart from variational distributions in \cref{eq:q_omega_u},
we assign an additional variational distribution on $\vv$, namely,
\begin{equation*}
q(\vv) = \mathcal{N}(\vmu^{(v)}, \mat{\Sigma}^{(v)}).
\end{equation*}
Due to the conjugate nature of GPs, we can get the approximated posterior of $\vf_{\bot}$,
\begin{equation*}
  q(\vf_{\bot}) = \int p(\vf_{\bot} \mid \vv) q(\vv) \dif \vv = \mathcal{N}(\vmu^{(f_{\bot})}, \mat{\Sigma}^{(f_{\bot})}),
\end{equation*}
where
\begin{align*}
  \vmu^{(f_{\bot})} &= \mC_{MH} \mC^{-1}_{HH} \vmu^{(v)}, \\
  \mat{\Sigma}^{(f_{\bot})} &= \tilde{\mK} + \mC_{MH} \mC_{HH}^{-1} ( \mat{\Sigma}^{(v)} - \mC_{HH} ) \mC_{HH}^{-1} \mC_{HM}, \\
  \mC_{MH} &= \mK_{MH} - \mK_{MB} \mK_{BB}^{-1} \mK_{BH}, \\
  \mC_{HH} &= \mK_{HH} - \mK_{HB} \mK^{-1}_{BB} \mK_{BH},
\end{align*}
with $\mK_{BH} = k(\mB, \mH)$.
Then we can derive the ELBO as
\begin{equation}\label{eq:solve-elbo}
  \begin{multlined}[.4\textwidth]
    \log p_{\theta}(\vx_{\Omega}, \mZ) \geq \E_{q(\vecs{\omega}) q(\vu)) q(\vf_{\bot})} \log p(\vx_{\Omega} \mid \vecs{\omega}, \vf) - \\
    \KL( q(\vu) q(\vv) q(\vecs{\omega}) \lVert p(\vu) p(\vv) p(\vecs{\omega}) ) + \log p(\mZ).
  \end{multlined}
\end{equation}
Note that the biggest difference between \cref{eq:svi-elbo1} and \cref{eq:solve-elbo}
is that, in \cref{eq:svi-elbo1} we take expectation
to $p(\vf \mid \vu)$, which is a prior, while in \cref{eq:solve-elbo},
we take expectation to $q(\vf_{\bot})$, which is a learnable posterior
and leads to more flexible learning processes.
Similarly, we can derive the closed-form updates for the PG variable,
\begin{equation*}
c_n = \sqrt{ \mu^{(f), 2}_{n} + \sigma^{(f_{\bot})}_{n, n} + \vecs{\kappa}_n^{(u), \intercal} \mat{\Sigma}^{(u)} \vecs{\kappa}_n^{(u)}},
\end{equation*}
where $\mu^{(f)}_n = \mu^{(f_{\bot})}_{n} + \vecs{\kappa}^{(u), \intercal}_{n} \vmu^{(u)}$ and $\sigma^{(f_{\bot})}_{n, n}$ is the $n$-th diagonal element of $\mat{\Sigma}^{(f_{\bot})}$.
The NG for $q(\vu)$ and $q(\vv)$ can also be derived.
\begin{align*}
  \tilde{\triangledown}_{\vecs{\eta}^{(i)}_{1}} &= \sum_{n=1}^N (\chi_{n} - \theta_{n} \vecs{\kappa}^{(j), \intercal}_{n} \vmu_n^{(j)}) \vecs{\kappa}^{(i)}_{n} - \vecs{\eta}^{(i)}_{1},  \\
  \tilde{\triangledown}_{\vecs{\eta}^{(i)}_{2}} &= - \vecs{\eta}^{(i)}_{2} - \frac{1}{2}(\mK_{II}^{-1} + \sum_{n=1}^N \theta_{n} \vecs{\kappa}^{(i)}_{n} \vecs{\kappa}^{(i), \intercal}_{n} ),
\end{align*}
where $i$ can be replaced by $u$ or $v$, $j=v$ if $i=u$ (vice versa), and $I = B$ if $i = u$ or $H$ if $i = v$. Moreover, $\vecs{\kappa}^{(v)}_{n}$ is defined as the $n$-th row of $\mC_{MH} \mC^{-1}_{HH}$.
As in \cref{sec:pg-augment},
the latent factors $\mZ$ and inducing inputs $\mB, \mH$ is then optimized
by maximizing the ELBO \cref{eq:solve-elbo}.
Detailed derivations and the whole algorithm are presented in \cref{sec:app-solve}.

\subsubsection{Complexity analysis}

Despite a more structured representation of variational approximations, SOLVE has additional
computational benefits due to the decoupled inducing points.
Suppose we choose $2p$ inducing points, computing \cref{eq:svi-elbo1}
requires $\mathcal{O}(4 N p^{2} + 8 p^{3})$ time complexity.
In SOLVE, supposing we decouple $2p$ points into two sets of $p$ points,
the complexity of optimizing \cref{eq:solve-elbo} reduces to $\mathcal{O}(2 N p^{2} + 2 p^{3})$.
Both approaches allow mini-batch training, which makes our model scalable to large datasets.

\section{Related Work}
\label{sec:related}

Traditional TDs usually rely on specific multi-linear contraction rules,
such as CP \citep{hitchcock1927expression}, Tucker \citep{tucker1966some}, tensor train/ring \citep[TT/TR,][]{oseledets2011tensor,zhao2019learning}
and many variants \citep{kolda2009tensor,cichocki2016tensor}.
While these models mainly focus on continuous cases,
binary and count data have also been considered.
\citet{chi2012tensors} proposed non-negative CP built on Poisson distribution.
Then, the Bayesian version of Poisson CP was established \citep{schein2015bayesian,schein2016bayesian} using gamma–Poisson conjugacy.
Concurrently, \citet{rai2014scalable,rai2015scalable} proposed Bayesian CP for binary and count data
using PG augmentation (PGA).
\citet{tao2023scalable} extended it to the TR format.
Recently,
\citet{wang2020learning} proposed a low-rank Bernoulli model based
on CP decomposition.
\citet{lee2021beyond} established a TD generated by summating a series of signs.
Generalized CP \citep[GCP,][]{hong2020generalized} summarized
diverse types of distributions and loss functions learned via
gradient-based methods.
\citet{soulat2021probabilistic} proposed a Bayesian
GCP using PGA to deal with NB
distribution for spiking count data.
All of these methods are based on multi-linear structures and may
lack flexibility for complex datasets.

\input{bin_completion.tex}

To enhance flexibility,
many non-linear TDs have been proposed.
In particular, \citet{pmlr-v5-chu09a,xu2012infinite,zhe2015scalable,zhe2016distributed}
proposed GP tensor factorizations (GPTFs) that use GPs to replace
multi-linear contractions. To deal with binary data, a Probit transform was adopted.
However, their model is unable to deal with count data.
Several following-up works adopt Gamma distribution
and Hawkes process to predict the happening time of each event \citep{zhe2018stochastic,pan2020scalable,wang2022nonparametric}.
Our model differs from these works in several ways.
Firstly, we adopt PGA to get a unified framework for both binary and count tensors.
Secondly, we derived efficient natural gradient updates.
Finally, a more efficient covariance approximation scheme is established for our model.
Recently, \citet{ibrahim2023under} proposed a TD for count data using
neural networks, which is out of the scope of this paper
as we use GPs. Moreover, it requires side information for each
mode, which is not always available.

In the context of GPs, handling non-conjugate models and establishing
efficient approximations are important topics.
To approximate large covariance matrices,
sparse variational Gaussian process \citep[SVGP,][]{titsias2009variational}
was proposed to variationally learn inducing points.
SVGP can be scaled to large datasets using stochastic optimization \citep{hensman2013gaussian}
or distributed learning \citep{gal2014distributed}.
Based on the idea of SVGP,
\citet{hensman2015scalable} proposed scalable GPs for binary classification.
\citet{wenzel2019efficient} proposed to use PG augmentation for GPs,
which yields conjugate models and fast NG updates.
Recently, \citet{shi2020sparse} proposed sparse orthogonal variational Gaussian process
(SOLVE-GP), which was shown to be more efficient in learning sparse GPs.
However, SOLVE-GP was not designed for binary and count data
and the authors did not utilize NG updates.
Therefore, our model is not only contributing to the TD community,
but also non-conjugate GPs in more general applications.

\section{Experiments}
\label{sec:experiment}

In this section, we present empirical evaluations of the proposed model.
All experiments are conducted on a workstation with
an Intel Xeon Silver 4316 CPU@2.30GHz, 512GB RAM and
NVIDIA RTX A6000 GPUs.
More details about experimental settings and results are
shown in \cref{sec:app-exp}.
The code is based on PyTorch \citep{paszke2019pytorch} and available at \url{https://github.com/taozerui/gptd}

\subsection{Binary Tensor Completion}
\label{sec:exp-bin}

\input{count_completion.tex}

\subsubsection{Datasets}

We test our model on three binary tensor datasets:
(1) Digg \citep{xu2012infinite}, an order-3 tensor of shape 581 \(\times\) 124 \(\times\) 48,
extracted from the \texttt{digg.com} social news website,
describing interactions among \emph{news} \(\times\) \emph{keyword} \(\times\) \emph{topic}.
It has 0.024\% non-zero entries.
(2) Enron \citep{xu2012infinite}, an order-3 tensor of shape 203 \(\times\) 203 \(\times\) 200,
storing records of an email system (\emph{sender} \(\times\) \emph{receiver} \(\times\) \emph{time}) with 0.01\% non-zero entries.
(3) DBLP \citep{zhe2016distributed}, an order-3 tensor of shape 10$k$ \(\times\) 200 \(\times\) 10$k$,
extracted from the DBLP database, depicting relationships among
\emph{author} \(\times\) \emph{conference} \(\times\) \emph{keyword}
with 0.001\% non-zero entries.
For Digg and Enron,
we randomly sample an equal number of zero entries to obtain a balanced dataset.
For DBLP,
the same train/test split with \citet{zhe2016distributed} is adopted.
For binary datasets, we evaluate the area under the ROC curve (AUC) and
the negative log-likelihood (NLL) of estimated Bernoulli distributions.
We report the mean and standard deviation of 5-fold cross-validation.

\subsubsection{Baselines}

We compare with five models:
(1) GCP \citep{hong2020generalized}, a generalized CP designed for diverse types
of data distributions and loss functions using gradient-based optimization.
(2) BCP \citep{wang2020learning}, a binary CPD with ALS-based algorithms.
(3) SBTR \citep{tao2023scalable}, a scalable Bayesian tensor ring that uses PGA to handle binary data,
which can be regarded as a TR version of \citet{rai2014scalable}.
(4) GPTF \citep{zhe2016distributed}, the GP tensor factorization that uses the Probit likelihood \cref{eq:gptf-origin} for binary data.
(5) CoSTCo \citep{liu2019costco}, a nonlinear TD uses convolutional neural networks for learn latent mappings.
Among the baselines, (1-3) are traditional multi-linear TDs and (4-5) are non-linear ones.
Note that CoSTCo was originally designed for continuous data, but can be easily fitted to binary domains (\cref{subsec:app-binary-tc}).

\subsubsection{Settings}

For baseline models, we mainly adopt their default settings.
All stochastic methods are optimized using batch size 128.
Moreover,
gradient-based models are optimized using Adam with
a learning rate chosen from $\{\num{3e-3}, \num{1e-3}, \num{3e-4}, \num{1e-4} \}$,
except GCP, whose default optimizer is L-BFGS.
We test all methods with different tensor ranks
ranging from \{ 3, 5, 10 \}.
For GP-based methods, we use 100 inducing points and
RBF kernel with bandwidth $1.0$,
consistent with previous work \citep{zhe2016distributed,zhe2018stochastic}.
Note that, for ENTED, the inducing points number is
50 + 50 for $\vu$ and $\vv$, respectively.

\subsubsection{Results}

\begin{figure}[h]
    \centering
    \begin{subfigure}{0.48\linewidth}
    \includegraphics[width=\linewidth]{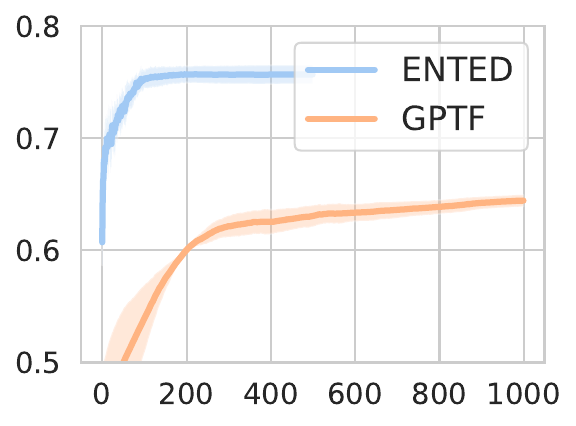}
    \caption{Digg}
    \end{subfigure}
    \hfill
    \begin{subfigure}{0.48\linewidth}
    \includegraphics[width=\linewidth]{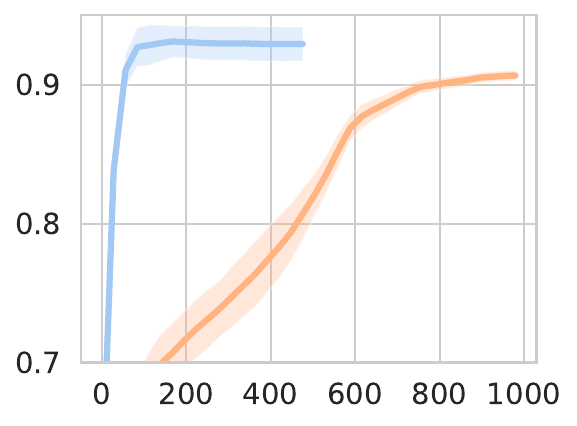}
    \caption{Enron}
    \end{subfigure}
    \caption{Convergence results. The x-axes are epochs and y-axes are AUC values.}
    \label{fig:converge}
\end{figure}

The completion results are shown in \cref{tab:completion-bin}.
For DBLP, the results of BCP are not available, due to its limited scalability.
Our model consistently outperforms competing models.
In particular, we observe that GPTF and our model perform much better
than other baselines, which shows the effectiveness of adopting non-linear
mappings.
Although CoSTCo is built upon non-linear CNNs, it performs poorly on these tasks.
We hypothesize two reasons.
When a binary tensor is generated from low-rank signals
after non-linear transforms, \eg, logistic transform,
the probability is not necessarily low-rank \citep{lee2021beyond}.
Thus, it is more reasonable to factorize the natural parameter rather than original binary observations.
Moreover, due to its highly unconstrained structures, CoSTCo easily overfits for sparse tensors.
Our model further outperforms GPTF, since we adopt more efficient covariance approximation
and stochastic NGs.
Moreover, due to the use of NGs,
our model converges much faster than GPTF.
The learning processes of rank 3 are illustrated
in \cref{fig:converge}.

\subsection{Count Tensor Completion}

\subsubsection{Datasets}

We evaluate the proposed model on three count tensors.
(1) JHU \citep{dong2020interactive}, an order-4 tensor of shape 51 \(\times\) 3 \(\times\) 48 \(\times\) 8, recording the Covid patient claims collected by JHU.
The maximum count is 5182.
The data is fully observed and
we use 20\% observations to predict the rest entries.
(2) Article \citep{zhe2018stochastic}, an order-3 tensor of shape 5 \(\times\) 1895 \(\times\) 2987, extracted from the DeskDrop dataset, recording user
operations to articles.
There are 50938 entries observed, and the maximum count is 76.
(3) EMS \citep{zhe2018stochastic}, an order-2 tensor recording the Emergency Medical Service (EMS) calls in Montgomery County, PA. The data shape is 72 \(\times\) 69, corresponding to
\emph{EMS title} \(\times\) \emph{township}.
There are 2494 observe entries and the maximum count is 545.
We evaluate our model using the relative root mean square error (RMSE),
mean absolute percentage error (MAPE), and negative log-likelihood (NLL).
The definitions are shown in \cref{subsec:app-count-tc}.

\subsubsection{Baselines}

We compare with six baselines.
Apart from (1) GCP, we also compare with:
(2) NCPD \citep{chi2012tensors}, a non-negative CP adopting Poisson likelihood.
(3) BPCP \citep{schein2015bayesian}, a Bayesian Poisson factorization with CP format.
(4) VB-GCP \cite{soulat2021probabilistic}, a Bayesian version of GCP learned via variational inference.
(5) GPTF \citep{zhe2016distributed}, a continuous GPTF using Gaussian likelihood.
(6) MDTF \citep{fan2022multimode}, a non-linear TD using neural networks to transform tensor factors.
Similarly, (1-4) are multi-linear TDs and (5-6) are non-linear models.
The settings are similar with \cref{sec:exp-bin}.
Therefore, we omit details here and present them in \cref{subsec:app-count-tc}.

\subsubsection{Results}

The RMSE and MAPE results
are shown in \cref{tab:completion-count}.
Our model outperforms baselines using Gaussian, Poisson, or NB distributions.
Compared with binary cases, the improvements over GPTF
are much more significant in count datasets,
which reveals the importance of choosing proper distributions.
Besides, our model outperforms VBGCP, which also employs
NB distribution, especially when the rank is small.
This indicates the advantage of using non-linear structures.
\cref{tab:count-nll} shows the NLL of several probabilistic baselines.
With a proper choice of the distribution, our model also achieves
much better distributional estimation, as opposed to GPTF with Gaussian likelihood.

\begin{table}[h]
  \centering
  \begin{tabular}{lccc}
    \toprule
    & \multicolumn{3}{c}{NLL\,$\downarrow$}           \\
    \cmidrule(r){2-4}
    \emph{JHU}   &  Rank $3$     & Rank $5$  & Rank $10$ \\
    \midrule
    GCP & INF &  INF & INF \\
    VBGCP & 62.1 $\pm$ 0.5 & 64.7 $\pm$ 0.6 & 69.0 $\pm$ 0.5  \\
    GPTF & 237. $\pm$ 40. & 216. $\pm$ 21. & 185. $\pm$ 25.  \\
    ENTED & \bftab 3.78 $\pm$ \bftab 0.01 & \bftab 3.74 $\pm$ \bftab 0.04 & \bftab 3.67 $\pm$ \bftab 0.07  \\
    \midrule
    \emph{Article}   &  \multicolumn{3}{c}{} \\
    \midrule
    GCP & 7.17 $\pm$ 0.17 & 7.10 $\pm$ 0.11 & 7.16 $\pm$ 0.17  \\
    VBGCP & 1.63 $\pm$ 0.01 & 1.63 $\pm$ 0.01 & 1.63 $\pm$ 0.01  \\
    GPTF & 1.71 $\pm$ 0.15 & 1.65 $\pm$ 0.12 & 1.70 $\pm$ 0.11  \\
    ENTED & \bftab 1.35 $\pm$ \bftab 0.01 & \bftab 1.35 $\pm$ \bftab 0.01 & \bftab 1.35 $\pm$ \bftab 0.01  \\
    \midrule
    \emph{EMS}   &  \multicolumn{3}{c}{} \\
    \midrule
    GCP & 11.8 $\pm$ 1.3 & 16.4 $\pm$ 1.6 & 27.9 $\pm$ 2.1  \\
    VBGCP & 23.2 $\pm$ 1.7 & 23.3 $\pm$ 1.7 & 22.5 $\pm$ 1.8  \\
    GPTF & 62.2 $\pm$ 20.4 & 54.3 $\pm$ 10.9 & 43.7 $\pm$ 11.2  \\
    ENTED & \bftab 2.95 $\pm$ \bftab 0.05 & \bftab 2.97 $\pm$ \bftab 0.08 & \bftab 2.91 $\pm$ \bftab 0.04  \\
    \bottomrule
  \end{tabular}
  \caption{NLL results of count tensor completion experiments. The NLL of GCP for the JHU dataset goes to infinity.}\label{tab:count-nll}
\end{table}

\subsection{Additional Results on Inducing Points}
\label{sec:exp-add}

Additional experiments are conducted to demonstrate the efficiency of our model.
In specific, different inducing point numbers are tested,
since it is essential for GP approximations.
Apart from GPTF, we also compare with the model presented in \cref{sec:pg-augment},
which is an extension of GPTF using PG augmentation and NG updates,
denoted as GPTF-PG.
To show the influence of the inducing points,
we fit the model of rank 10 on the \emph{Article} dataset,
varying inducing points number in $\{ 2^{3}, 2^{6}, 2^{7}, 2^{8}, 2^{9}, 2^{10}\}$.
Similarly, in ENTED, the inducing point numbers for $\vu$ and $\vv$ are equal.

\begin{figure}[h]
    \centering
    \begin{subfigure}{0.48\linewidth}
    \includegraphics[width=\linewidth]{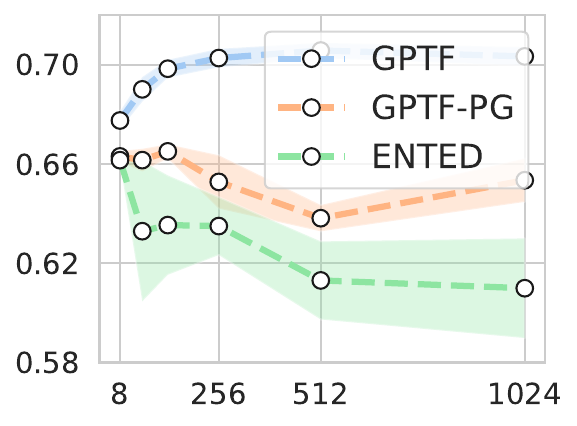}
    \caption{RMSE}\label{fig:inducing-rmse}
    \end{subfigure}
    \hfill
    \begin{subfigure}{0.48\linewidth}
    \includegraphics[width=\linewidth]{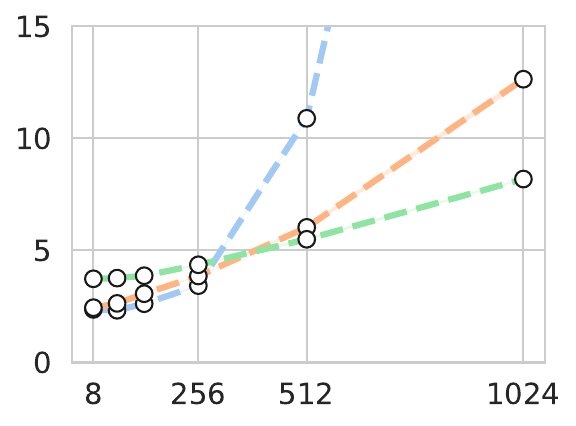}
    \caption{Computing Time}\label{fig:inducing-time}
    \end{subfigure}
    \caption{Results on different numbers of inducing points. The x-axes denote the number of inducing points. In subfigure (a), the y-axis is the RMSE and in subfigure (b), the y-axis is the computing time (in seconds) for each epoch.}
    \label{fig:inducing}
\end{figure}

\cref{fig:inducing-rmse} shows that our model consistently
outperforms GPTF and GPTF-PG.
For GPTF, more inducing points lead to worse performance, maybe due to over-fitting.
However, for GPTF-PG and ENTED,
the performance improves as the inducing points grow,
since they adopt more structured inference and NG.
\cref{fig:inducing-time} shows the computing time of one epoch.
Our model is slower than GPTF and GPTF-PG when the inducing points number is small
since we need to implement the NG updates manually.
Nevertheless, when the inducing points number becomes larger,
our model is notably faster than GPTF and GPTF-PG.
This reveals the computational advantage of our model in handling complex datasets,
where a large number of inducing points may be needed.

\section{Conclusions}

An efficient nonparametric tensor decomposition for binary and count data is presented.
By replacing traditional multi-linear products with non-linear Gaussian processes (GP), the model can capture more complex relations among different tensor modes.
By using the \pg augmentation, a conjugate model and natural gradient updates can be derived analytically. Moreover, we derived the sparse orthogonal variational inference framework to enable faster and more flexible covariance matrix approximation. Experiments are conducted on binary and count tensor completion tasks, to show the superior performance and computational benefits of our model.
In the future, it is of interest to extend our model in streaming data and continual learning settings.

\section*{Acknowledgments}

Zerui Tao was supported by the RIKEN Junior Research Associate Program.
This work was supported by the
JSPS KAKENHI Grant Numbers JP20H04249, JP23H03419.

\bibliography{aaai24}

\input{appendix}

\end{document}

%% file: math_commands.tex
\usepackage{amsmath,amsfonts,bm}

\def\eqref#1{equation~\ref{#1}}

\def\1{\bm{1}}

\def\vzero{{\bm{0}}}

\def\vmu{{\bm{\mu}}}

\def\vc{{\bm{c}}}

\def\vf{{\bm{f}}}

\def\vm{{\bm{m}}}

\def\vu{{\bm{u}}}
\def\vv{{\bm{v}}}

\def\vx{{\bm{x}}}

\def\vz{{\bm{z}}}

\newcommand{\vecs}[1]{{\bm{#1}}}

\def\mB{{\bm{B}}}
\def\mC{{\bm{C}}}

\def\mH{{\bm{H}}}
\def\mI{{\bm{I}}}

\def\mK{{\bm{K}}}
\def\mL{{\bm{L}}}
\def\mM{{\bm{M}}}

\def\mX{{\bm{X}}}

\def\mZ{{\bm{Z}}}

\newcommand{\mat}[1]{{\bm{#1}}}

\DeclareMathAlphabet{\mathsfit}{\encodingdefault}{\sfdefault}{m}{sl}
\SetMathAlphabet{\mathsfit}{bold}{\encodingdefault}{\sfdefault}{bx}{n}
\newcommand{\tens}[1]{\bm{\mathsfit{#1}}}

\def\tX{{\tens{X}}}

\newcommand{\E}{\mathbb{E}}

\newcommand{\KL}{D_{\mathrm{KL}}}



%% file: bin_completion.tex
\begin{table*}[ht]
  \centering
  \begin{tabular}{lccc|ccc}
    \toprule
    & \multicolumn{3}{c|}{AUC\,$\uparrow$} &  \multicolumn{3}{c}{NLL\,$\downarrow$}          \\
    \cmidrule(r){2-7}
    \emph{Digg}     &  Rank $3$     & Rank $5$  &  Rank $10$  &  Rank $3$     & Rank $5$   & Rank $10$ \\
    \midrule
    GCP & 0.566 $\pm$ 0.034 & 0.555 $\pm$ 0.026 & 0.539 $\pm$ 0.045 & 5.356 $\pm$ 0.302 & 6.747 $\pm$ 1.011 & 8.696 $\pm$ 1.874  \\
    BCP & 0.564 $\pm$ 0.016 & 0.566 $\pm$ 0.031 & 0.547 $\pm$ 0.024 & 0.688 $\pm$ 0.003 & 0.690 $\pm$ 0.003 & 0.689 $\pm$ 0.001  \\
    SBTR & 0.698 $\pm$ 0.028 & 0.694 $\pm$ 0.039 & 0.728 $\pm$ 0.024 & 0.647 $\pm$ 0.013 & 0.646 $\pm$ 0.012 & 0.632 $\pm$ 0.009  \\
    GPTF & 0.629 $\pm$ 0.025 & 0.655 $\pm$ 0.028 & 0.737 $\pm$ 0.017 & 0.663 $\pm$ 0.013 & 0.662 $\pm$ 0.010 & 0.611 $\pm$ 0.019  \\
    CoSTCo & 0.594 $\pm$ 0.033 & 0.597 $\pm$ 0.047 & 0.620 $\pm$ 0.076 & 0.693 $\pm$ 0.015 & 0.693 $\pm$ 0.014 & 0.679 $\pm$ 0.123  \\
    ENTED & \bftab 0.720 $\pm$ \bftab 0.039 & \bftab 0.743 $\pm$ \bftab 0.039 & \bftab 0.767 $\pm$ \bftab 0.032 & \bftab 0.622 $\pm$ \bftab 0.041 & \bftab 0.599 $\pm$ \bftab 0.030 & \bftab 0.577 $\pm$ \bftab 0.030  \\
    \midrule
    \emph{Enron}     &  \multicolumn{6}{c}{} \\
    \midrule
    GCP & 0.848 $\pm$ 0.010 & 0.852 $\pm$ 0.022 & 0.847 $\pm$ 0.028 & 3.501 $\pm$ 0.611 & 3.541 $\pm$ 0.948 & 3.608 $\pm$ 1.070  \\
    BCP & 0.664 $\pm$ 0.055 & 0.651 $\pm$ 0.060 & 0.652 $\pm$ 0.059 & 0.599 $\pm$ 0.008 & 0.590 $\pm$ 0.010 & 0.590 $\pm$ 0.008  \\
    SBTR & 0.899 $\pm$ 0.031 & 0.905 $\pm$ 0.019 & 0.896 $\pm$ 0.010 & 0.487 $\pm$ 0.039 & 0.470 $\pm$ 0.036 & 0.476 $\pm$ 0.022  \\
    GPTF & 0.921 $\pm$ 0.015 & 0.928 $\pm$ 0.019 & \bftab 0.950 $\pm$ \bftab 0.013 & 0.398 $\pm$ 0.053 & 0.366 $\pm$ 0.043 & 0.349 $\pm$ 0.038  \\
    CoSTCo & 0.549 $\pm$ 0.072 & 0.602 $\pm$ 0.045 & 0.838 $\pm$ 0.114 & 0.693 $\pm$ 0.010 & 0.693 $\pm$ 0.014 & 0.485 $\pm$ 0.477  \\
    ENTED & \bftab 0.926 $\pm$ \bftab 0.012 & \bftab 0.938 $\pm$ \bftab 0.004 & \bftab 0.950 $\pm$ \bftab 0.013 & \bftab 0.381 $\pm$ \bftab 0.047 & \bftab 0.359 $\pm$ \bftab 0.025 & \bftab 0.309 $\pm$ \bftab 0.063  \\
    \midrule
    \emph{DBLP}     &  \multicolumn{6}{c}{} \\
    \midrule
    GCP & 0.926 $\pm$ 0.003 & 0.941 $\pm$ 0.002 & 0.950 $\pm$ 0.002 & 0.836 $\pm$ 0.004 & 0.804 $\pm$ 0.004 & 0.775 $\pm$ 0.006  \\
    SBTR & 0.897 $\pm$ 0.008 & 0.915 $\pm$ 0.004 & 0.954 $\pm$ 0.000 & 0.460 $\pm$ 0.029 & 0.440 $\pm$ 0.010 & 0.317 $\pm$ 0.005  \\
    GPTF & 0.942 $\pm$ 0.003 & 0.945 $\pm$ 0.003 & 0.954 $\pm$ 0.002 & 0.384 $\pm$ 0.015 & 0.363 $\pm$ 0.013 & 0.339 $\pm$ 0.010  \\
    CoSTCo & 0.892 $\pm$ 0.006 & 0.904 $\pm$ 0.006 & 0.907 $\pm$ 0.001 & 0.440 $\pm$ 0.502 & 0.403 $\pm$ 0.518 & 0.381 $\pm$ 0.513  \\
    ENTED & \bftab 0.950 $\pm$ \bftab 0.003 & \bftab 0.959 $\pm$ \bftab 0.003 & \bftab 0.962 $\pm$ \bftab 0.002 & \bftab 0.286 $\pm$ \bftab 0.009 & \bftab 0.263 $\pm$ \bftab 0.010 & \bftab 0.250 $\pm$ \bftab 0.007  \\
    \bottomrule
  \end{tabular}
  \caption{Binary tensor completion.}\label{tab:completion-bin}
  \vspace{-1em}
\end{table*}

%% file: count_completion.tex
\begin{table*}[ht]
  \centering
  \begin{tabular}{lccc|ccc}
    \toprule
    & \multicolumn{3}{c|}{RMSE\,$\downarrow$} &  \multicolumn{3}{c}{MAPE\,$\downarrow$}          \\
    \cmidrule(r){2-7}
    \emph{JHU}     &  Rank $3$     & Rank $5$  & Rank $10$  &  Rank $3$  & Rank $5$  & Rank $10$ \\
    \midrule
    GCP & 0.847 $\pm$ 0.017 & 0.857 $\pm$ 0.008 & 0.873 $\pm$ 0.013 & 0.666 $\pm$ 0.003 & 0.668 $\pm$ 0.002 & 0.677 $\pm$ 0.001  \\
    NCPD & 0.856 $\pm$ 0.015 & 0.861 $\pm$ 0.014 & 0.876 $\pm$ 0.013 & 0.666 $\pm$ 0.004 & 0.668 $\pm$ 0.005 & 0.674 $\pm$ 0.003  \\
    BPCP & 0.852 $\pm$ 0.015 & 0.862 $\pm$ 0.014 & 0.881 $\pm$ 0.005 & 0.669 $\pm$ 0.002 & 0.671 $\pm$ 0.004 & 0.678 $\pm$ 0.001  \\
    VBGCP & 0.681 $\pm$ 0.255 & 0.484 $\pm$ 0.104 & 0.567 $\pm$ 0.255 & 0.365 $\pm$ 0.055 & 0.299 $\pm$ 0.013 & \bftab 0.286 $\pm$ \bftab 0.024  \\
    GPTF & 0.505 $\pm$ 0.019 & 0.508 $\pm$ 0.018 & 0.546 $\pm$ 0.014 & 0.496 $\pm$ 0.089 & 0.403 $\pm$ 0.014 & 0.391 $\pm$ 0.011  \\
    MDTF & 0.554 $\pm$ 0.116 & 0.529 $\pm$ 0.098 & 0.522 $\pm$ 0.047 & 0.715 $\pm$ 0.041 & 1.030 $\pm$ 0.218 & 0.578 $\pm$ 0.074  \\
    ENTED & \bftab 0.438 $\pm$ \bftab 0.026 & \bftab 0.426 $\pm$ \bftab 0.026 & \bftab 0.406 $\pm$ \bftab 0.042 & \bftab 0.313 $\pm$ \bftab 0.005 & \bftab 0.296 $\pm$ \bftab 0.011 & 0.292 $\pm$ 0.037  \\
    \midrule
    \emph{Article}     &  \multicolumn{6}{c}{} \\
    \midrule
    GCP & 0.946 $\pm$ 0.003 & 0.939 $\pm$ 0.004 & 0.929 $\pm$ 0.005 & 0.511 $\pm$ 0.002 & 0.506 $\pm$ 0.002 & 0.499 $\pm$ 0.002  \\
    NCPD & 0.937 $\pm$ 0.004 & 0.931 $\pm$ 0.006 & 0.923 $\pm$ 0.009 & 0.507 $\pm$ 0.003 & 0.502 $\pm$ 0.003 & 0.498 $\pm$ 0.004  \\
    BPCP & 0.945 $\pm$ 0.003 & 0.940 $\pm$ 0.005 & 0.938 $\pm$ 0.006 & 0.513 $\pm$ 0.002 & 0.510 $\pm$ 0.003 & 0.509 $\pm$ 0.003  \\
    VBGCP & 0.805 $\pm$ 0.026 & 0.804 $\pm$ 0.026 & 0.804 $\pm$ 0.026 & 0.342 $\pm$ 0.006 & 0.341 $\pm$ 0.006 & 0.340 $\pm$ 0.006  \\
    GPTF & 0.630 $\pm$ 0.026 & 0.629 $\pm$ 0.028 & 0.655 $\pm$ 0.027 & 0.190 $\pm$ 0.005 & 0.197 $\pm$ 0.001 & 0.206 $\pm$ 0.003  \\
    MDTF & 0.749 $\pm$ 0.030 & 0.783 $\pm$ 0.027 & 0.851 $\pm$ 0.079 & 0.216 $\pm$ 0.011 & 0.224 $\pm$ 0.011 & 0.253 $\pm$ 0.016  \\
    ENTED & \bftab 0.620 $\pm$ \bftab 0.020 & \bftab 0.628 $\pm$ \bftab 0.026 & \bftab 0.636 $\pm$ \bftab 0.024 & \bftab 0.164 $\pm$ \bftab 0.003 & \bftab 0.170 $\pm$ \bftab 0.004 & \bftab 0.173 $\pm$ \bftab 0.002  \\
    \midrule
    \emph{EMS}     &  \multicolumn{6}{c}{} \\
    \midrule
    GCP & 0.728 $\pm$ 0.049 & 0.852 $\pm$ 0.047 & 0.955 $\pm$ 0.008 & 0.431 $\pm$ 0.013 & 0.474 $\pm$ 0.006 & 0.562 $\pm$ 0.009  \\
    NCPD & 0.835 $\pm$ 0.029 & 0.904 $\pm$ 0.014 & 0.968 $\pm$ 0.003 & 0.437 $\pm$ 0.009 & 0.478 $\pm$ 0.005 & 0.565 $\pm$ 0.014  \\
    BPCP & 0.722 $\pm$ 0.082 & 0.869 $\pm$ 0.008 & 0.936 $\pm$ 0.005 & 0.429 $\pm$ 0.007 & 0.474 $\pm$ 0.013 & 0.542 $\pm$ 0.003  \\
    VBGCP & 0.345 $\pm$ 0.064 & 0.343 $\pm$ 0.062 & 0.321 $\pm$ 0.048 & 0.428 $\pm$ 0.042 & 0.423 $\pm$ 0.040 & 0.398 $\pm$ 0.032  \\
    GPTF & 0.530 $\pm$ 0.062 & 0.460 $\pm$ 0.048 & 0.403 $\pm$ 0.060 & 0.957 $\pm$ 0.149 & 0.637 $\pm$ 0.026 & 0.458 $\pm$ 0.063  \\
    MDTF & 0.377 $\pm$ 0.055 & 0.381 $\pm$ 0.057 & 0.387 $\pm$ 0.055 & 0.667 $\pm$ 0.138 & 0.738 $\pm$ 0.119 & 0.809 $\pm$ 0.181  \\
    ENTED & \bftab 0.305 $\pm$ \bftab 0.049 & \bftab 0.319 $\pm$ \bftab 0.056 & \bftab 0.301 $\pm$ \bftab 0.045 & \bftab 0.399 $\pm$ \bftab 0.058 & \bftab 0.414 $\pm$ \bftab 0.102 & \bftab 0.355 $\pm$ \bftab 0.027  \\
    \bottomrule
  \end{tabular}
  \caption{Count tensor completion.}\label{tab:completion-count}
  \vspace{-1em}
\end{table*}

%% file: appendix.tex
\begin{appendix}

\onecolumn

\section{Gaussian Process Tensor Factorization for Binary Data}
\label{sec:app-gptf}

In this section, we present an extension of the Gaussian process tensor factorization \citep[GPTF,][]{zhe2016distributed}.
This model mainly differs from the original one in the optimization mechanism.
In \citet[]{zhe2016distributed}, the authors derived an ELBO that can be computed
in a \emph{distributed} way. However, the objective function is not factorized over
samples and cannot be optimized using \emph{stochastic} methods.
Here, we instead to use stochastic variational inference (SVI) which is more scalable
for many real-world applications.
Specifically, for binary data, we adopt the Probit likelihood,
\begin{equation}\label{eq:app-gptf-probit}
p(x_{\ind_{n}} \mid \omega_{n}) = \mathcal{B}(\Phi(\omega_{n})) = \Phi(\omega_n)^{x_{\ind_{n}}} (1 - \Phi(\omega_n))^{(1 - x_{\ind_{n}})}, \quad \forall n = 1, \dots, N,
\end{equation}
where $\omega$ is an auxiliary variable and
$\Phi(\cdot)$ is the cumulative distribution function (CDF) of the standard Gaussian distribution.
The auxiliary variable $\omega_{n}$ is constructed as follows,
\begin{equation*}
  \omega_n \mid f_n \sim \mathcal{N}(\omega_n \mid f_n, 1).
\end{equation*}
Then, we assign GPTF on the latent variable $\vf$.
We denote all latent factors associated with index $\ind$ as
$\vm_{\ind} = [\vz^{(1)}_{i_{1}}, \dots, \vz^{(D)}_{i_{D}}] \in \mathbb{R}^{DR}$,
where $\vz^{(d)}_{i_{d}} \in \mathbb{R}^{R}$ is the $i_{d}$-th row of $\mZ^{(d)} $.
Then the linear contraction forms of traditional TDs can be replaced by a GP, namely $\vf \sim \mathcal{GP}(0, k(\vm, \cdot))$. For finite sample case, we have
\begin{equation}\label{eq:app-gptf-gp}
  p(\vf) = \mathcal{N}(\vf \mid \vecs{0}, k(\mM_{\Omega}, \mM_{\Omega})),
\end{equation}
where $k(\cdot, \cdot)$ is a kernel function, $\Omega = \{ \ind_1, \dots, \ind_{N}  \}$ denotes all observed indices and $\mM_{\Omega} \in \mathbb{R}^{N \times DR}$ is concatenated by latent factors associating with all observed entries, \ie, $\{ \vm_{\ind}: \ind \in \Omega \}$.
Finally, the joint pdf becomes,
\begin{equation}\label{eq:app-gptf-joint}
  p(\vx_{\Omega}, \vf, \vecs{\omega}, \mZ^{(1)}, \dots, \mZ^{(D)}) = \prod_{n=1}^{N} p(x_{\ind_{n}} \mid \omega_{n}) \cdot p(\omega_{n} \mid f_{n}) \cdot p(\vf) \cdot \prod_{d=1}^{D} p(\mZ^{(d)}),
\end{equation}
where $p(\mZ^{(d)})$ is the prior distribution of latent factors
and we can simply set $p(\mZ^{(d)})= \mathcal{N}(\vect(\mZ^{(d)}) \mid \vzero, \mI)$.
To learn the model,
we aim to maximize the joint likelihood $\log p(\vx_{\Omega}, \mZ)$ by marginalizing
out $\vf, \vecs{\omega}$ in \cref{eq:app-gptf-joint}.

However, exactly computing the joint likelihood is computational infeasible.
Here, we adopt the framework of
sparse variational Gaussian process \citep[SVGP,][]{titsias2009variational,hensman2013gaussian,hensman2015scalable}.
In specific, we adopt $p \ll N$ inducing
inputs $\mB \in \mathbb{R}^{p \times DR}$ and
corresponding inducing points $\vecs{u} \in \mathbb{R}^{p}$, satisfying,
\begin{equation*}
p(\vf \mid \vu) = \mathcal{N}(\mK_{MB} \mK_{BB}^{-1} \vu, \tilde{\mK}), \quad p(\vu) = \mathcal{N}(\vzero, \mK_{BB}),
\end{equation*}
where
\[  \mK_{MB} = k(\mM_{\Omega}, \mB), \quad \mK_{BB} = k(\mB, \mB), \quad \mK_{MM} = k(\mM_{\Omega}, \mM_{\Omega}), \quad \tilde{\mK} = \mK_{MM} - \mK_{MB} \mK_{BB}^{-1} \mK_{BM}.  \]
The joint distribution becomes,
\begin{equation}\label{eq:app-gptf-joint2}
  p(\vx_{\Omega}, \vf, \vecs{\omega}, \mZ, \vu) = \prod_{n=1}^{N} p(x_{\ind_{n}} \mid \omega_{n}) \cdot p(\omega_{n} \mid f_{n}) \cdot p(\vf \mid \vu) \cdot p(\vu) \cdot \prod_{d=1}^{D} p(\mZ^{(d)}),
\end{equation}
where we denote $\mZ = \{ \mZ^{(d)} \}_{d=1}^{D}$ for simplicity.
To get a tractable lower bound for the model evidence (ELBO),
we firstly marginalize out $\vf$ as follows,
\begin{align}
  \log p(\vecs{\omega} \mid \vu, \mZ) &= \log \int p(\vecs{\omega} \mid \vf) p(\vf \mid \vu) \dif \vf \nonumber \\
  &\geq \E_{p(\vf \mid \vu)} \log p(\vecs{\omega} \mid \vf) \nonumber \\
  &= \sum_{n =1}^{N} \log \mathcal{N}(\omega_{n} \mid \vecs{\kappa}^{(u), \intercal}_{n} \vu, 1) - \frac{1}{2} \tilde{k}_{n, n}, \label{eq:app-gptf-pomegau}
\end{align}
where $\vecs{\kappa}^{(u)}_{n}$ is the $n$-th row of $\mK_{MB} \mK_{BB}^{-1}$ and $\tilde{k}_{n, n}$ is the $(n, n)$-th element of $\tilde{\mK}$.
Then, in order to marginalize out $\vu$, we introduce a variational approximation,
\begin{equation}\label{eq:app-qu}
q(\vu) = \mathcal{N}(\vmu^{(u)}, \mat{\Sigma}^{(u)}).
\end{equation}
The variational lower bound (ELBO) is obtained by injecting \cref{eq:app-qu} into \cref{eq:app-gptf-pomegau},
\begin{align}
  \log p(\vecs{\omega} \mid \mZ) &= \log \int p(\vecs{\omega} \mid \vu, \mZ) p(\vu) \dif \vu \nonumber \\
  &\geq \E_{q(\vu)} [ \log p(\vecs{\omega} \mid \vu, \mZ) ] - \KL (q(\vu) \lVert p(\vu)) \nonumber \\
  &= \sum_{n=1}^N \left\{ \log \mathcal{N}(\omega_{n} \mid \vecs{\kappa}_{n}^{(u), \intercal} \vmu^{(u)}, 1) - \frac{1}{2} \tilde{k}_{n, n} - \frac{1}{2} \tr(\mat{\Sigma}^{(u)} \mat{\Lambda}_n) \right\} - \KL(q(\vu) \lVert p(\vu)), \label{eq:app-gptf-l2}
\end{align}
where $\mat{\Lambda}_{n} = \vecs{\kappa}^{(u)}_n \vecs{\kappa}^{(u), \intercal}_{n}$.
Then, from \cref{eq:app-gptf-l2}, we have
\begin{align}
  &p(\vx_{\Omega} \mid \mZ) \nonumber \\
  =& \int p(\vx_{\Omega} \mid \vecs{\omega}) p(\vecs{\omega} \mid \mZ) \dif \vecs{\omega} \nonumber \\
  =& \left( \prod_{n=1}^N \int p(x_{\ind_{n}} \mid \omega_n) \mathcal{N}(\omega_n \mid \vecs{\kappa}^{(u), \intercal}_{n} \vmu^{(u)}, 1) \dif \omega_{n} \right) \cdot \exp \left( \sum_{n=1}^N \left\{ - \frac{1}{2} \tilde{k}_{n, n} - \frac{1}{2} \tr(\mat{\Sigma}^{(u)} \mat{\Lambda}_{n}) \right\} - \KL(q(\vu) \lVert p(\vu))  \right). \label{eq:app-gptf-pxz}
\end{align}
Finally, plugging \cref{eq:app-gptf-pxz} into \cref{eq:app-gptf-joint2}, we have
\begin{equation}
\label{eq:app-gptf-l3}
\begin{multlined}[.8\linewidth]
  \log (\vx_{\Omega}, \mZ) \geq \sum_{n=1}^{N} \left\{ x_{\ind_{n}} \log \Phi\left( \frac{\vecs{\kappa}^{(u), \intercal}_{n} \vmu^{(u)}}{\sqrt{2}} \right) + (1 - x_{\ind_{n}})\log\left( 1 - \Phi\left( \frac{\vecs{\kappa}^{(u), \intercal}_{n} \vmu^{(u)}}{\sqrt{2}} \right)  \right) - \frac{1}{2} \tilde{k}_{n, n} - \frac{1}{2} \tr(\mat{\Sigma}^{(u)} \mat{\Lambda}_{n}) \right\} \\
    - \frac{1}{2} \log \lvert \mK_{BB} \mat{\Sigma}^{(u), -1} \lvert - \frac{1}{2} \tr( \mat{\Sigma}^{(u)} \mK^{-1}_{BB}) - \frac{1}{2} \vmu^{(u), \intercal} \mK^{-1}_{BB} \vmu^{(u)} - \sum_{d=1}^{D} \lVert \mZ^{(d)} \lVert^{2}_{F}.
\end{multlined}
\end{equation}
The variational distribution $q(\vu) = \mathcal{N}(\vmu^{(u)}, \mat{\Sigma}^{(u)})$,
the latent factors $\mZ$ and inducing inputs $\mB$ are optimized by maximizing
the ELBO \cref{eq:app-gptf-l3}.
In practice, we parameterize the variational distribution as
$q(\vu) = \mathcal{N}(\vmu^{(u)}, \mL^{(u)} \mL^{(u), \intercal})$,
where $\mL^{(u)}$ is a lower triangle matrix,
and use reparameterization trick to compute the expectations.
This approximation reduces complexity to $\mathcal{O}(p^{3} + Np^{2})$.
More importantly, objective \cref{eq:app-gptf-l2} is factorized over observations,
so that stochastic optimization is possible and $N$ can be replaced by mini-batch sizes.

\section{Proposed Model}

\subsection{Nonparametric Tensor Decomposition with \pg Augmentation}
\label{sec:app-pga}

\subsubsection{\pg Augmentation}

Firstly, we introduced basic backgrounds of \pg (PG) distribution \citep{polson2013bayesian}
and how to use PG variables to augment GPTF.

\begin{definition}[\pg distribution]
  Suppose $\omega$ follows the \pg distribution, $p(\omega) = PG(b, c)$, then
  \[ \omega \stackrel{D}{=} \frac{1}{2 \pi^{2}} \sum_{k=1}^{\infty} \frac{g_k}{(k - 1 / 2)^2 + c^2 / (4 \pi^{2})}, \]
  where $g_k \sim Ga(b, 1)$ independently $\forall k$, and $\stackrel{D}{=}$ means equality in distribution.
\end{definition}
Here we list several properties of PG distribution, which are essential for our derivation.
\begin{enumerate}
  \item For $\omega \sim PG(b, c)$, we have
        \begin{equation}
        \label{eq:app-pg-pdf}
          PG(\omega \mid b, c) = \cosh^b\left( \frac{c}{2} \right) \exp \left(- \frac{c^{2}}{2} \omega \right) PG(\omega \mid b, 0).
        \end{equation}
  \item The first-order moment (expectation) of a PG variable is
        \begin{equation}
        \label{eq:app-pg-exp}
          \E_{PG(\omega \mid b, c)} [ \omega ] = \frac{b}{2c} \tanh \left( \frac{c}{2} \right).
        \end{equation}
  \item Suppose $\omega \sim PG(b, 0)$, we have
    \begin{equation}
    \label{eq:app-pg-exp-exp}
    \frac{\exp(t)^a}{(1 + \exp(t))^b} = 2^{-b} \exp((a - b/2) t) \int_0^{\infty} \exp(- \omega t^2 / 2) p(\omega) \dif \omega.
    \end{equation}
\end{enumerate}

Due to \cref{eq:app-pg-exp-exp}, we can recover Bernoulli or NB distribution
by marginalizing out $\omega$ in \cref{eq:pga}.

\subsubsection{Evidence Lower Bound}
To obtain the lower bound for the joint distribution \cref{eq:pg-joint},
we firstly marginalize out the latent variable $\vf$,
\begin{align}
\log p(\vx_{\Omega} \mid \vecs{\omega}, \vu, \mZ) &= \log \int p(\vx_{\Omega} \mid \vecs{\omega}, \vf) p(\vf \mid \vu, \mZ) \dif \vf \nonumber \\
  &\geq \E_{p(\vf \mid \vu, \mZ)} \log p(\vx \mid \vecs{\omega}, \vf) \nonumber \\
  &\propto \E_{p(\vf \mid \vu, \mZ)} \sum_{n=1}^{N} \chi_{n} f_{n} - \frac{1}{2} f_n^2 \omega_{n} \nonumber \\
  &= \sum_{n=1}^N \chi_{n} \vecs{\kappa}^{(u), \intercal}_{n} \vu - \frac{1}{2} \omega_{n} ((\vecs{\kappa}^{(u), \intercal}_{n} \vu)^{2} + \tilde{k}_{n, n}), \label{eq:app-pg-px}
\end{align}
where $\vecs{\kappa}_{n}^{(u)}$ is the $n$-th row of $\mK_{MB}\mK_{BB}^{-1}$ (For notation consistency, we treat it as a \emph{column} vectors) and $\tilde{k}_{n, n}$ is the $n$-th diagonal element of $\tilde{\mK}$ defined in \cref{eq:gp-fu}.
Similar to \cref{sec:app-gptf},
we introduce a variational distribution $q(\vu, \vecs{\omega}) = q(\vu) q(\vecs{\omega})$,
where
\begin{equation*}
q(\vu) = \mathcal{N}(\vu \mid \vmu^{(u)}, \vecs{\Sigma}^{(u)}), \quad q(\vecs{\omega}) = \prod_{n=1}^N PG(\omega_n \mid b, c_{n}).
\end{equation*}
Then, we plugging \cref{eq:app-pg-px} into \cref{eq:pg-joint} and use the variational distributions to marginalize out the latent variables
\begin{align}
  &\log p(\vx, \mZ) \nonumber \\
  \geq& \E_{q(\vu) q(\vecs{\omega})} \log p(\vx \mid \vecs{\omega}, \vu, \mZ) - \KL(q(\vu, \vecs{\omega}) \lVert p(\vu) p(\vecs{\omega})) + \sum_{d=1}^{D} \log p(\mZ^{(d)}) \nonumber \\
  =& \begin{multlined}[.7\linewidth]
      \frac{1}{2} \sum_{n=1}^{N} \left\{ 2 \chi_{n} \vecs{\kappa}^{(u), \intercal}_{n} \vmu^{(u)} - \theta_{n} ( \tilde{k}_{n, n} + \vecs{\kappa}^{(u), \intercal}_n \mat{\Sigma}^{(u)} \vecs{\kappa}_n^{(u)} + \vmu^{(u), \intercal} \vecs{\kappa}_n^{(u)} \vecs{\kappa}^{(u), \intercal}_n \vmu^{(u)}) + c_n^2 \theta_{n} - 2 b \log \cosh(c_n / 2) \right\} \\
      - \frac{1}{2} \log \lvert \mK_{BB} \mat{\Sigma}^{(u), -1} \lvert - \frac{1}{2} \tr( \mat{\Sigma}^{(u)} \mK^{-1}_{BB}) - \frac{1}{2} \vmu^{(u), \intercal} \mK^{-1}_{BB} \vmu^{(u)} - \frac{1}{2} \sum_{d=1}^{D} \lVert \mZ^{(d)} \lVert^{2}_{F},
  \end{multlined} \label{eq:app-pg-elbo}
\end{align}
where $\theta_{n} = \frac{b}{2 c_{n}} \tanh(c_{n} / 2)$.
This ELBO is obtained by simply taking the tractable expectation of the log-likelihood term and computing the KL divergences.
In specific, the KL divergence $\KL(q(\omega_{n}) \lVert p(\omega_{n}))$ can be drived as followls.
Using \cref{eq:app-pg-pdf}, we have
\begin{equation*}
p(\omega_n) = PG(b, 0), \quad
q(\omega_n) = PG(b, c_n) = \cosh^{b}(c_n / 2) \exp\left(- \frac{c^2_n}{2} \omega_n\right) PG(\omega_n \mid b, 0).
\end{equation*}
Then, by adopting \cref{eq:app-pg-exp},
the KL divergence becomes,
\begin{align*}
\KL(q(\omega) \lVert p(\omega)) &= \E_{q(\omega)}[ \log q(\omega) - \log p(\omega) ] \\
  &= \E_{q(\omega)}\log \left( \cosh^{b}(c / 2) \exp\left(- \frac{c^2}{2} \omega\right) PG(\omega \mid b, 0) \right) - \E_{q(\omega)} \log PG(\omega \mid b, 0) \\
  &= \log \cosh^{b}(c / 2) - \frac{bc}{4} \tanh(c / 2) + \cancel{ \E_{q(\omega)} PG(\omega \mid b, 0) - \E_{q(\omega)} \log PG(\omega \mid b, 0) }\\
  &= b \log \cosh(c / 2) - \frac{bc}{4} \tanh(c / 2),
\end{align*}
where we omit the subscript $n$ for simplicity.

\subsection{Stochastic Variational Inference with Natural Gradients}
\label{sec:app-elbo-ng}

In this subsection, we present the full derivation of
the natural gradient (NG) updates. The derivation of this subsection follows \citet{wenzel2019efficient}.
The gradient of the ELBO \cref{eq:app-pg-elbo} \wrt the local parameter $c_n$ is,
\begin{align*}
  &\frac{\partial \log p(\vx, \mZ)}{\partial c_{n}} \\
  =&  \frac{\partial}{\partial c_{n}} \left[ - \frac{b}{4c_{n}} \tanh\left( \frac{c_{n}}{2} \right) A_{n} + \frac{bc_{n}}{4} \tanh\left( \frac{c_{n}}{2} \right) - \log \cosh^b \left(\frac{c_{n}}{2}\right) \right] \\
  =& \frac{A_{n} b}{4 c_{n}^{2}} \tanh \left(\frac{c_{n}}{2}\right) - \frac{b}{2} \cdot \frac{A_{n}}{4c_{n}} \left(1 - \tanh^2\left(\frac{c_{n}}{2}\right)\right) + \frac{b}{4} \tanh\left(\frac{c_{n}}{2}\right) + \frac{1}{2} \cdot \frac{bc_{n}}{4}\left(1 - \tanh^{2}\left(\frac{c_{n}}{2}\right)\right) - \frac{b}{2} \tanh\left(\frac{c_{n}}{2}\right) \\
  =& \left(\frac{A_{n} b}{4 c_{n}^{2}} - \frac{b}{4}\right) \tanh\left(\frac{c_{n}}{2}\right) - \frac{b}{2}\left(\frac{A_{n}}{4c_{n}} - \frac{bc_{n}}{4}\right)\left( 1- \tanh^2\left(\frac{c_{n}}{2}\right) \right) \\
  =& \left(\frac{A_{n} b}{4 c_{n}^2} - \frac{b}{4}\right) \left(\tanh\left(\frac{c_{n}}{2}\right) - \frac{c_{n}}{2}\left(1 - \tanh^2\left(\frac{c_{n}}{2}\right) \right) \right),
\end{align*}
where
\[ A_{n} =
\tilde{k}_{n, n} + \vecs{\kappa}^{(u), \intercal}_n \mat{\Sigma}^{(u)} \vecs{\kappa}_n^{(u)} + \vmu^{(u), \intercal} \vecs{\kappa}_n^{(u)} \vecs{\kappa}^{(u), \intercal}_n \vmu^{(u)}.
\]
To get closed-form updates, we let the above gradient to zero.
Then second term equals to zero when $c_n = 0$, hence is neglected.
Letting$\displaystyle \frac{A_{n} b}{4 c_{n}^2} - \frac{b}{4} = 0$, since $b \neq 0$, we have,
\begin{equation*}
  c_n = \sqrt{\tilde{k}_{n, n} + \vecs{\kappa}^{(u), \intercal}_n \mat{\Sigma}^{(u)} \vecs{\kappa}_n^{(u)} + \vmu^{(u), \intercal} \vecs{\kappa}_n^{(u)} \vecs{\kappa}^{(u), \intercal}_n \vmu^{(u)}}.
\end{equation*}

Then, we derive the natural gradients.
The gradient of \cref{eq:app-pg-elbo} is
\begin{align*}
\frac{\partial \mathcal{L}}{\partial \vmu^{(u)}} &= - \mK_{BB}^{-1} \vmu^{(u)} + \sum_{n=1}^N ( \chi_{n} - \theta_n \vecs{\kappa}^{(u), \intercal}_n \vmu^{(u)})\vecs{\kappa}^{(u)}_n, \\
\frac{\partial \mathcal{L}}{\partial \mat{\Sigma}^{(u)}} &= \frac{1}{2} \left( \mat{\Sigma}^{(u), -1} - \mK_{BB}^{-1} - \sum_{n=1}^N \theta_{n} \vecs{\kappa}^{(u)}_{n} \vecs{\kappa}^{(u), \intercal}_{n} \right).
\end{align*}
The natural parameters are
\begin{equation*}
  \vecs{\eta}_1 = \mat{\Sigma}^{-1} \vmu, \quad \vecs{\eta}_2 = - \frac{1}{2} \mat{\Sigma}^{-1},
\end{equation*}
where we omit the superscript $(u)$ for simplicity.
And the NG is defined as
\begin{equation*}
\tilde{\triangledown}_{(\vecs{\eta}_{1}, \vecs{\eta}_{2})} = (\triangledown_{\vmu} \mathcal{L} - 2 \triangledown_{\mat{\Sigma}} \mathcal{L} \cdot \vmu, \triangledown_{\mat{\Sigma}} \mathcal{L} ).
\end{equation*}
Therefore, we have
\begin{equation*}
\tilde{\triangledown}_{\vecs{\eta}^{(u)}_{1}} = \sum_{n=1}^N \chi_{n} \vecs{\kappa}^{(u)}_{n} - \vecs{\eta}^{(u)}_{1},  \quad \tilde{\triangledown}_{\vecs{\eta}^{(u)}_{2}} = - \vecs{\eta}^{(u)}_{2} - \frac{1}{2}(\mK_{BB}^{-1} + \sum_{n=1}^N \theta_{n} \vecs{\kappa}^{(u)}_{n} \vecs{\kappa}^{(u), \intercal}_{n} ),
\end{equation*}

\subsection{Efficient Orthogonally Decoupled Approximation}
\label{sec:app-solve}

To compute the ELBO \cref{eq:solve-elbo}, we firstly compute the expectation of the log-likelihood,
\begin{align}
  &\E_{q(\vu) q(\vf_{\bot}) q(\vecs{\omega})} \log p(\vx \mid \vecs{\omega}, \vf)  \nonumber \\
  =&\sum_{n=1}^{N} \E_{q(\vu) q(\vf_{\bot}) q(\vecs{\omega})} [ \chi_{n} (f_{\bot, n} + \vecs{\kappa}_{n}^{(u), \intercal} \vu) - \frac{1}{2} \omega_n ( f^2_{\bot, n} + 2 f_{\bot, n} \vecs{\kappa}^{(u), \intercal}_{n} \vu + (\vecs{\kappa}^{(u), \intercal}_{n} \vu)^{2} )]  \nonumber \\
  =& \frac{1}{2} \sum_{n=1}^N \E_{q(\vu) q(\vecs{\omega})} [ 2 \chi_{n} \mu^{f_{\bot}}_{n} + 2 \chi_{n} \vecs{\kappa}_{n}^{(u), \intercal} \vu - \omega_n (\mu^{(f_{\bot}), 2}_{n} + \sigma^{(f_{\bot})}_{n, n}) - 2 \omega_n \mu^{(f_{\bot})}_n \vecs{\kappa}^{(u), \intercal}_{n} \vu - \omega_n(\vecs{\kappa}^{(u), \intercal}_{n} \vu)^{2} ]  \nonumber \\
  =& \begin{multlined}[.7\linewidth]
     \frac{1}{2} \sum_{n=1}^N \E_{q(\vecs{\omega})} [ 2 \chi_{n} \mu^{f_{\bot}}_{n} + 2 \chi_{n} \vecs{\kappa}_{n}^{(u), \intercal} \vmu^{(u)} - \omega_n (\mu^{(f_{\bot}), 2}_{n} + \sigma^{(f_{\bot})}_{n, n}) - 2 \omega_n \mu^{(f_{\bot})}_n \vecs{\kappa}^{(u), \intercal}_{n} \vmu^{(u)} \\
  - \omega_n(\vecs{\kappa}^{(u), \intercal}_n \mat{\Sigma}^{(u)} \vecs{\kappa}_n^{(u)} + \vmu^{(u), \intercal} \vecs{\kappa}_n^{(u)} \vecs{\kappa}^{(u), \intercal}_n \vmu^{(u)}) ]
  \end{multlined} \nonumber \\
  =& \begin{multlined}[.7\linewidth]
     \frac{1}{2} \sum_{n=1}^N 2 \chi_{n} \mu^{f_{\bot}}_{n} + 2 \chi_{n} \vecs{\kappa}_{n}^{(u), \intercal} \vmu^{(u)} - \theta_n (\mu^{(f_{\bot}), 2}_{n} + \sigma^{(f_{\bot})}_{n, n}) - 2 \theta_n \mu^{(f_{\bot})}_n \vecs{\kappa}^{(u), \intercal}_{n} \vmu^{(u)}  \\
  - \theta_n(\vecs{\kappa}^{(u), \intercal}_n \mat{\Sigma}^{(u)} \vecs{\kappa}_n^{(u)} + \vmu^{(u), \intercal} \vecs{\kappa}_n^{(u)} \vecs{\kappa}^{(u), \intercal}_n \vmu^{(u)}),
  \end{multlined} \label{eq:app-solve-ll}
\end{align}
where $\displaystyle \theta_n = \frac{b}{2 c_{n}} \tanh( c_n / 2 )$ and $\sigma^{(f_{\bot})}_{n, n}$ is the $n$-th diagonal element of $\mat{\Sigma}^{(f_{\bot})}$.
The KL divergence terms are similar with previous section,
\begin{align}
\KL(q(\vecs{\omega}) \lVert p(\vecs{\omega})) &= \sum_{n=1}^{N} \log \cosh^{b}(c_n / 2) - \frac{bc_{n}}{4} \tanh(\frac{c_{n}}{2}), \label{eq:app-solve-kl-omega} \\
\KL(q(\vu) \lVert p(\vu)) &= \frac{1}{2} \log \lvert \mK_{BB} \mat{\Sigma}^{(u), -1} \lvert + \frac{1}{2} \tr( \mat{\Sigma}^{(u)} \mK^{-1}_{BB}) + \frac{1}{2} \vmu^{(u), \intercal} \mK^{-1}_{BB} \vmu^{(u)}, \label{eq:app-solve-kl-u} \\
\KL(q(\vv) \lVert p(\vv)) &= \frac{1}{2} \log \lvert \mK_{HH} \mat{\Sigma}^{(v), -1} \lvert + \frac{1}{2} \tr( \mat{\Sigma}^{(v)} \mK^{-1}_{HH}) + \frac{1}{2} \vmu^{(v), \intercal} \mK^{-1}_{HH} \vmu^{(v)}.\label{eq:app-solve-kl-v}
\end{align}
Plugging \cref{eq:app-solve-ll,eq:app-solve-kl-omega,eq:app-solve-kl-u,eq:app-solve-kl-v} into
the \cref{eq:solve-elbo}, we can get the ELBO.
Using similar derivation with \cref{sec:app-elbo-ng}, we can derive the update rule for local parameters,
\begin{align}
c_n &= \sqrt{ \mu^{(f_{\bot}), 2}_{n} + \sigma^{(f_{\bot})}_{n, n} + 2  \mu^{(f_{\bot})}_n \vecs{\kappa}^{(u), \intercal}_{n} \vmu^{(u)} + \vecs{\kappa}_n^{(u)} \mat{\Sigma}^{(u)} \vecs{\kappa}_n^{(u), \intercal} + \vmu^{(u), \intercal} \vecs{\kappa}_n^{(u), \intercal} \vecs{\kappa}^{(u)}_n \vmu^{(u)} } \nonumber \\
  &= \sqrt{ \mu^{(f), 2}_{n} + \sigma^{(f_{\bot})}_{n, n} + \vecs{\kappa}_n^{(u), \intercal} \mat{\Sigma}^{(u)} \vecs{\kappa}_n^{(u)}}, \label{eq:app-update-omega}
\end{align}
where $\mu^{(f)}_n = \mu^{(f_{\bot})}_{n} + \vecs{\kappa}^{(u), \intercal}_{n} \vmu^{(u)}$.
Finally, we derive the natural gradients, as follows,
\begin{align*}
\frac{\partial \mathcal{L}}{\partial \vmu^{(u)}} &= - \mK_{BB}^{-1} \vmu^{(u)} + \sum_{n=1}^N \left( \chi_{n} - \theta_n \mu^{(f_\bot)}_n - \theta_n \vecs{\kappa}^{(u), \intercal}_n \vmu^{(u)} \right) \vecs{\kappa}^{(u)}_n, \\
\frac{\partial \mathcal{L}}{\partial \mat{\Sigma}^{(u)}} &= \frac{1}{2} \left( \mat{\Sigma}^{(u), -1} - \mK_{BB}^{-1} - \sum_{n=1}^N \theta_{n} \vecs{\kappa}^{(u)}_{n} \vecs{\kappa}^{(u), \intercal}_{n} \right).
\end{align*}
And
\begin{align*}
\frac{\partial \mathcal{L}}{\partial \vmu^{(v)}} &= - \mK_{HH}^{-1} \vmu^{(v)} + \sum_{n=1}^N \left( \chi_{n} - \theta_n \vecs{\kappa}^{(u), \intercal} \vmu^{(u)} - \theta_n \vecs{\kappa}^{(v), \intercal}_n \vmu^{(v)} \right) \vecs{\kappa}^{(v)}_n, \\
\frac{\partial \mathcal{L}}{\partial \mat{\Sigma}^{(v)}} &= \frac{1}{2} \left( \mat{\Sigma}^{(v), -1} - \mK_{HH}^{-1} - \sum_{n=1}^N \theta_{n} \vecs{\kappa}^{(v)}_{n} \vecs{\kappa}^{(v), \intercal}_{n} \right),
\end{align*}
where $\vecs{\kappa}^{(v)}_n$ is the $n$-th row of $\mC_{MH} \mC^{-1}_{HH}$.
Then, similar with \cref{sec:app-elbo-ng}, we have
\begin{equation}\label{eq:app-solve-ng-u}
\tilde{\triangledown}_{\vecs{\eta}^{(u)}_{1}} = \sum_{n=1}^N (\chi_{n} - \theta_{n} \vecs{\kappa}^{(v), \intercal}_n \vmu^{(v)}) \vecs{\kappa}^{(u)}_{n} - \vecs{\eta}^{(u)}_{1},  \quad \tilde{\triangledown}_{\vecs{\eta}^{(u)}_{2}} = - \vecs{\eta}^{(u)}_{2} - \frac{1}{2}(\mK_{BB}^{-1} + \sum_{n=1}^N \theta_{n} \vecs{\kappa}^{(u)}_{n} \vecs{\kappa}^{(u), \intercal}_{n} ),
\end{equation}
and
\begin{equation}\label{eq:app-solve-ng-v}
\tilde{\triangledown}_{\vecs{\eta}^{(v)}_{1}} = \sum_{n=1}^N (\chi_{n} - \theta_{n} \vecs{\kappa}^{(u), \intercal}_{n} \vmu^{(u)}) \vecs{\kappa}^{(v)}_{n} - \vecs{\eta}^{(v)}_{1},  \quad \tilde{\triangledown}_{\vecs{\eta}^{(v)}_{2}} = - \vecs{\eta}^{(v)}_{2} - \frac{1}{2}(\mK_{HH}^{-1} + \sum_{n=1}^N \theta_{n} \vecs{\kappa}^{(v)}_{n} \vecs{\kappa}^{(v), \intercal}_{n} ).
\end{equation}

The overall algorithm is summarized in \cref{alg:algorithm}.

\begin{algorithm}[h]
\caption{Efficient Nonparametric Tensor Decomposition}
\label{alg:algorithm}
\textbf{Input}: Observed order-$D$ tensor $\tX$, observed indices $\Omega$. \\
\textbf{Output}: Latent factors $\mZ$, inducing inputs $\mB, \mH \in \mathbb{R}^{p \times RD}$ and variational distributions $q(\vecs{\omega}, q(\vu), q(\vv)$. \\
\textbf{Hyper-parameter}: Rank $R$, inducing point number $p$, initial learning rate $\lambda$, number of successes $\zeta$.
\begin{algorithmic}[1] 
\STATE Randomly initialize $\mZ, \mH, \mB$ and $\vmu^{(i)}, \mat{\Sigma}^{(i)}$ for $i = u, v$.
\WHILE{not converge}
\STATE Sample a minibatch of entries $\vx_{\mathcal{S}}$.
\STATE \COMMENT{Update variational distribution $q(\vecs{\omega})$}
\STATE Update the parameters in $q(\vecs{\omega})$ by \cref{eq:app-update-omega}.
\STATE \COMMENT{Update variational distribution $q(\vu)$}
\STATE Compute natural parameters $\vecs{\eta}^{(u)}_1 = \mat{\Sigma}^{(u), -1} \vmu^{(u)}$ and $\vecs{\eta}^{(u)}_2 = - \frac{1}{2} \mat{\Sigma}^{(u), -1}$.
\STATE Update $\vecs{\eta}^{(u)}_1$ and $\vecs{\eta}^{(u)}_2$ by gradients in \cref{eq:app-solve-ng-u}
\STATE Compute $\mat{\Sigma}^{(u)} = - \frac{1}{2} \vecs{\eta}^{(u), -1}_2$ and $\vmu^{(u)} = \vecs{\eta}^{(v)}_2 \mat{\Sigma}^{(u), -1}$.
\STATE \COMMENT{Update variational distribution $q(\vv)$}
\STATE Compute natural parameters $\vecs{\eta}^{(v)}_1 = \mat{\Sigma}^{(v), -1} \vmu^{(v)}$ and $\vecs{\eta}^{(v)}_2 = - \frac{1}{2} \mat{\Sigma}^{(v), -1}$.
\STATE Update $\vecs{\eta}^{(v)}_1$ and $\vecs{\eta}^{(v)}_2$ by gradients in \cref{eq:app-solve-ng-v}
\STATE Compute $\mat{\Sigma}^{(v)} = - \frac{1}{2} \vecs{\eta}^{(v), -1}_2$ and $\vmu^{(v)} = \vecs{\eta}^{(v)}_2 \mat{\Sigma}^{(v), -1}$.
\STATE \COMMENT{Update other parameters}
\STATE Compute the ELBO \cref{eq:solve-elbo} by plugging in \cref{eq:app-solve-ll,eq:app-solve-kl-omega,eq:app-solve-kl-u,eq:app-solve-kl-v}.
\STATE Update $\mZ, \mB, \mH$ by maximizing the ELBO using gradient ascent.
\ENDWHILE
\end{algorithmic}
\end{algorithm}

\section{Experiments}
\label{sec:app-exp}

\subsection{Binary Tensor completion}%
\label{subsec:app-binary-tc}

\paragraph{Baselines}

We compare with the following baselines.
\begin{enumerate}
\item GCP \citep{hong2020generalized},
      a generalized CPD designed for diverse types
      of data distributions and loss functions using gradient-based optimization.
      The model is provided in the Tensor Toolbox\footnote{\url{https://www.tensortoolbox.org/}} for \textsc{Matlab}.
\item BCP \citep{wang2020learning}, a binary CPD with ALS-based 
      algorithms. As a concequence, this model cannot scale to
      large tensors.
      The code is available at the repository\footnote{\url{https://github.com/Miaoyanwang/Binary-Tensor}}.
\item SBTR \citep{tao2023scalable}, a scalable Bayesian tensor ring
      that uses PGA to handle binary data, which can be regarded
      as a TR version of \citet{rai2014scalable}.
      This model is implemented based on PyTorch\footnote{\url{https://github.com/taozerui/scalable_btr}}.
\item GPTF \citep{zhe2016distributed},
      the GP tensor factorization that uses the Probit likelihood.
      This model is slightly different from \citet{zhe2016distributed},
      as we described in \cref{sec:app-gptf}.
      We implement this model using PyTorch.
\item CoSTCo \citep{liu2019costco},
      a nonlinear TD uses convolutional neural networks.
      We employ the official implementation\footnote{\url{https://github.com/USC-Melady/KDD19-CoSTCo}}.
      However, to deal with binary data, we add a sigmoid activation
      for output and optimize the binary cross entropy loss.
\end{enumerate}
Among the baselines, (1-3) are traditional multi-linear TDs and (4-5) are non-linear ones.
We run GCP, BCP on the CPU and run SBTR, GPTF, CoSTCo, ENTED
on GPUs.

\subsubsection{Settings}

For baseline models, we mainly adopt their default settings.
All stochastic methods are optimized using batch size 128.
Moreover,
gradient-based models are optimized using Adam with
learning rate chosen from $\{\num{3e-3}, \num{1e-3}, \num{3e-4}, \num{1e-4} \}$,
except GCP, whose default optimizer is L-BFGS.
We test all methods with different tensor ranks
ranging from \{ 3, 5, 10 \}.
For GP-based methods, we use 100 inducing points and
RBF kernel with bandwidth $1.0$,
which is consistent with previous works \citep{zhe2016distributed,zhe2018stochastic}.
Note that, for ENTED, the inducing points number is
50 + 50 for $\vu$ and $\vv$, respectively.

\subsection{Count Tensor completion}%
\label{subsec:app-count-tc}

\subsubsection{Baselines}

We compare with six baselines.
\begin{enumerate}
\item GCP \citep{hong2020generalized}.
      We choose the Poisson CP model.
\item NCPD \citep{chi2012tensors},
      a non-negative CP adopting Poisson likelihood.
      This model is also provided in the Tensor Toolbox
      for \textsc{Matlab}.
\item BPCP \citep{schein2015bayesian},
      a Bayesian Poisson factorization with CP format.
      The code is provided in the repository\footnote{\url{https://github.com/aschein/bptf}}.
\item VB-GCP \cite{soulat2021probabilistic},
      a Bayesian version of GCP learned via variational inference.
      This model also adopts NB distributions.
      The ELBO is optimized using ALS-like coordinate ascent
      variational inference (CAVI), which is not scalable to
      large tensors.
      We adopt the \textsc{Matlab} implemention provided in the repository\footnote{\url{https://github.com/hugosou/vbgcp}}.
\item GPTF \citep{zhe2016distributed},
      a continuous GPTF using Gaussian likelihood.
      This model is the same with the one in \cref{sec:app-gptf},
      except using Gaussian distribution.
\item MDTF \citep{fan2022multimode},
      a non-linear TD using neural networks to transform tensor factors. The code is provided in the repository\footnote{\url{https://github.com/jicongfan/Multi-Mode-Deep-Matrix-and-Tensor-Factorization}}.
\end{enumerate}
Similarly, (1-4) are multi-linear TDs and (5-6) are non-linear models.
In addition, (1-4) are designed for count tensor completion.
While (4-5) are initially based on Gaussian distribution,
we treat the count observations as coutinuous for these two models.
We run GCP, NCPD, BPCP, VB-GCP, MDTF on the CPU and run GPTF, ENTED
on GPUs.

\subsubsection{Settings}
The settings are the same with binary completion experiments.
For baseline models, we mainly use their default settings.
For GP-based models, including GPTF and ENTED, we set inducing points to 100 as before.
For our model, there is one hyper-parameter, \ie, the shape $\zeta$
of NB distribution, which is set to $20$ for all datasets.
For count datasets,
we evaluate our model using the relative root mean square error (RMSE),
mean absolute percentage error (MAPE) and negative log-likelihood (NLL), defined as follows,
\begin{equation*}
    \mathrm{RMSE} = \frac{\sqrt{\sum_{n=1}^N (x_n - \hat{x}_n)^2}}{\sqrt{\sum_{n=1}^N x^2_n}}, \quad
    \mathrm{MAPE} = \frac{1}{N} \sum_{n=1}^N \frac{| x_n - \hat{x}_n |}{| x_n + 1 |},
\end{equation*}
where $\hat{x}_n$ are estimates
We add 1 in the denominator of MAPE since the count may be zero.
Moreover, the NLL depends on different distributions the model
utilizes.

\end{appendix}